%% file: main.tex
\documentclass[10pt,twocolumn,letterpaper]{article}

\usepackage[pagenumbers]{cvpr} 

\input{preamble}

\title{ManiFPT: Defining and Analyzing Fingerprints of Generative Models}

\author{Hae Jin Song$^{1}$\thanks{Corresponding author: Hae Jin Song <haejinso@usc.edu>} \quad Mahyar Khayatkhoei$^{1}$  \quad Wael AbdAlmageed$^{1,2}$ \\
$^1$USC Information Sciences Institute \\
$^2$USC Ming Hsieh Department of Electrical and Computer Engineering \\
}

\begin{document}
\maketitle

\input{sec/00-abstract}



\input{tbls/fig-our-defn-fpts}   
\input{tbls/fig-tsne-gallery}  

\section{Introduction} \label{sec:intro}
\input{sec/01-intro}

\section{Related Work} \label{sec:rel-work}
\input{sec/02-rel-work}

\input{tbls/fig-our-attribution-workflow} 

\section{ManiFPT: Manifold-based Fingerprints of generative models} \label{sec:our-method}    

\input{sec/03-our-method}

\input{tbls/fig-viz-our-defns}  

\section{Experiments} \label{sec:exps} 
\input{sec/04-exps}

\section{Conclusion} \label{sec:conclusion}
\input{sec/05-conclusion}

{
    \small
    \bibliographystyle{ieee_fullname}
    \bibliography{main}
}

\input{sec/06-appendix}

\end{document}

%% file: preamble.tex
\usepackage[dvipsnames]{xcolor}

\usepackage{times}
\usepackage{epsfig}
\usepackage[utf8]{inputenc} 
\usepackage[T1]{fontenc}    
\usepackage{url}            
\usepackage{nicefrac}       
\usepackage{microtype}      
\usepackage{xcolor}         

\usepackage{caption}
\usepackage{subcaption}  
\usepackage{graphicx}
\graphicspath{{./figs/}}
\usepackage{float}       
\usepackage{adjustbox}   
\usepackage{makecell}    
\usepackage{arydshln}   

\definecolor{cvprblue}{rgb}{0.21,0.49,0.74}
\usepackage[pagebackref,breaklinks,colorlinks,citecolor=cvprblue]{hyperref}

\usepackage[capitalize]{cleveref}
\crefname{section}{Sec.}{Secs.}
\Crefname{section}{Section}{Sections}
\Crefname{table}{Table}{Tables}
\crefname{table}{Tab.}{Tabs.}

\usepackage{amsthm}
\theoremstyle{definition}
\newtheorem{definition}{Definition}[section]
\theoremstyle{remark}

\newcommand\xleftrightarrow[1]{%
  \mathbin{\ooalign{$\,\xrightarrow{#1}$\cr$\xleftarrow{\hphantom{#1}}\,$}}
}

\input{commands}
\newcommand\oursrgb{\textit{ManiFPT}_\textsubscript{RGB}}
\newcommand\oursfreq{\textit{ManiFPT}_\textsubscript{FREQ}}
\newcommand\ourssup{\textit{ManiFPT}_\textsubscript{SL}}
\newcommand\oursunsup{\textit{ManiFPT}_\textsubscript{SSL}}

\newcommand\Phigt{\mathcal{M}} 

\newcommand\phirgb{x_{\text{RGB}}}
\newcommand\phifreq{x_{\text{FREQ}}}
\newcommand\phisl{x_{\text{SL}}}
\newcommand\phissl{x_{\text{SSL}}}

\newcommand{\gmcifar}{GM-CIFAR10}
\newcommand{\gmceleba}{GM-CelebA}
\newcommand{\gmchq}{GM-CHQ}
\newcommand{\gmffhq}{GM-FFHQ}

%% file: commands.tex
\DeclareMathOperator*{\argmin}{\arg\!\min}

\usepackage{algpseudocode}
\usepackage{algorithm}
\usepackage{algorithmicx}

\algrenewcommand\algorithmicrequire{\textbf{Precondition:}}  
\algrenewcommand\algorithmicensure{\textbf{Postcondition:}}

\usepackage{xspace}
\makeatletter
\DeclareRobustCommand\onedot{\futurelet\@let@token\@onedot}
\def\@onedot{\ifx\@let@token.\else.\null\fi\xspace}

\def\eg{\emph{e.g}\onedot} 
\def\ie{\emph{i.e}\onedot}

\makeatletter

%% file: sec/00-abstract.tex
\begin{abstract}
Recent works have shown that generative models leave traces of their underlying generative process on the generated samples, broadly referred to as fingerprints of a generative model, and have studied their utility in detecting synthetic images from real ones. However, the extend to which these fingerprints can distinguish between various types of synthetic image and help identify the underlying generative process remain under-explored. In particular, the very definition of a fingerprint remains unclear, to our knowledge. To that end, in this work, we formalize the definition of artifact and fingerprint in generative models, propose an algorithm for computing them in practice, and finally study its effectiveness in distinguishing a large array of different generative models. We find that using our proposed definition can significantly improve the performance on the task of identifying the underlying generative process from samples (model attribution) compared to existing methods. Additionally, we study the structure of the fingerprints, and observe that it is very predictive of the effect of different design choices on the generative process.

\end{abstract}

%% file: tbls/fig-our-defn-fpts.tex
\begin{figure}[t]
    \centering
    \includegraphics[width=0.5\textwidth]{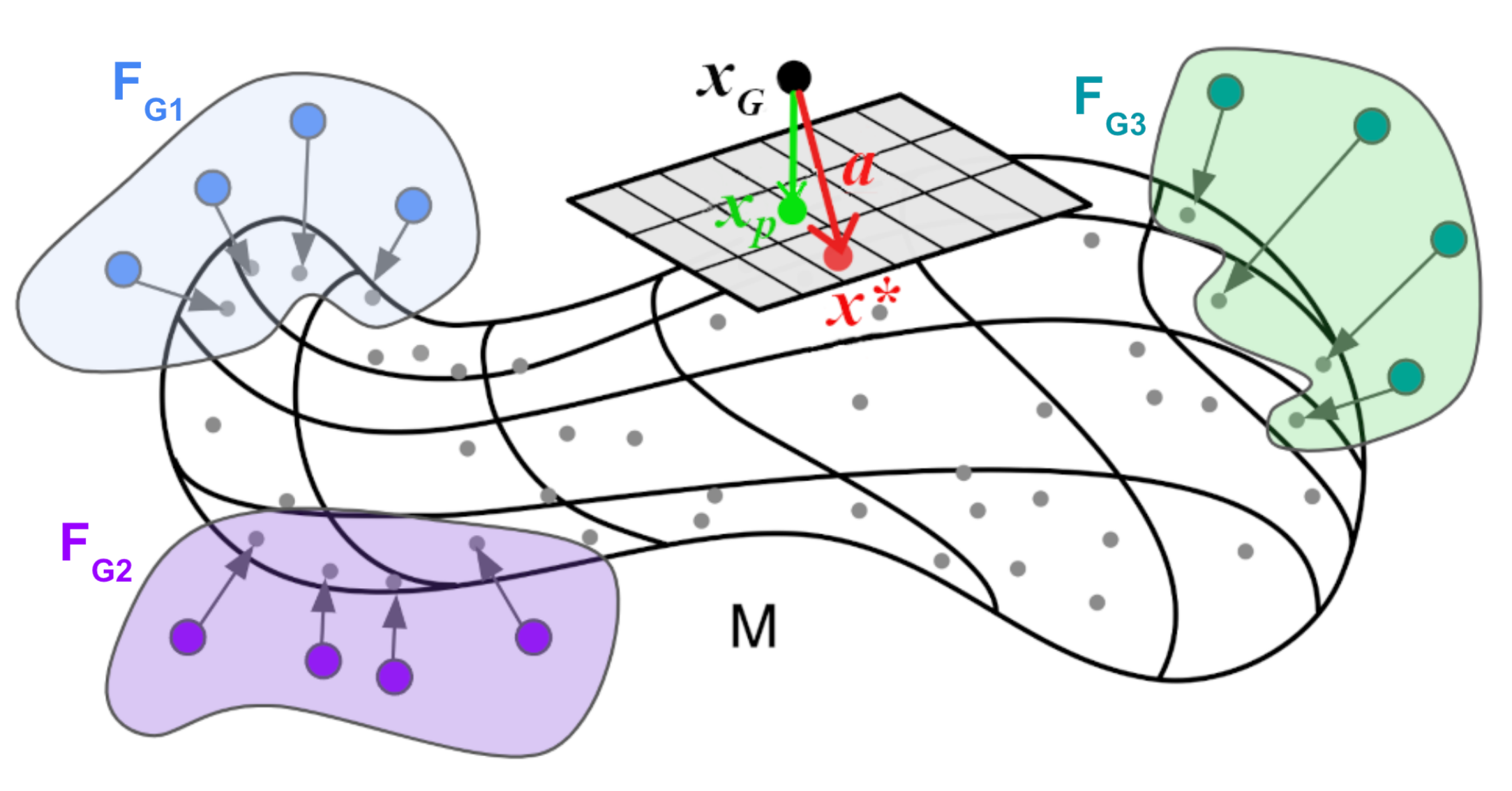} 
    \caption{\textbf{Our definition of artifacts and fingerprints of a generative model.}
    We estimate the true data manifold $\mathcal{M}$ using real samples and compute an artifact $a$ as the difference between a generated sample and its closest point in the real dataset. 
    We define the fingerprint F of a generative model as the set of all its artifacts. 
    \\    
    }
    \label{fig:our-defn-fpts}
    \vspace{-1.5em}
\end{figure}

%% file: tbls/fig-tsne-gallery.tex
\begin{figure*}[t]
    \centering
    \includegraphics[width=0.95\linewidth]{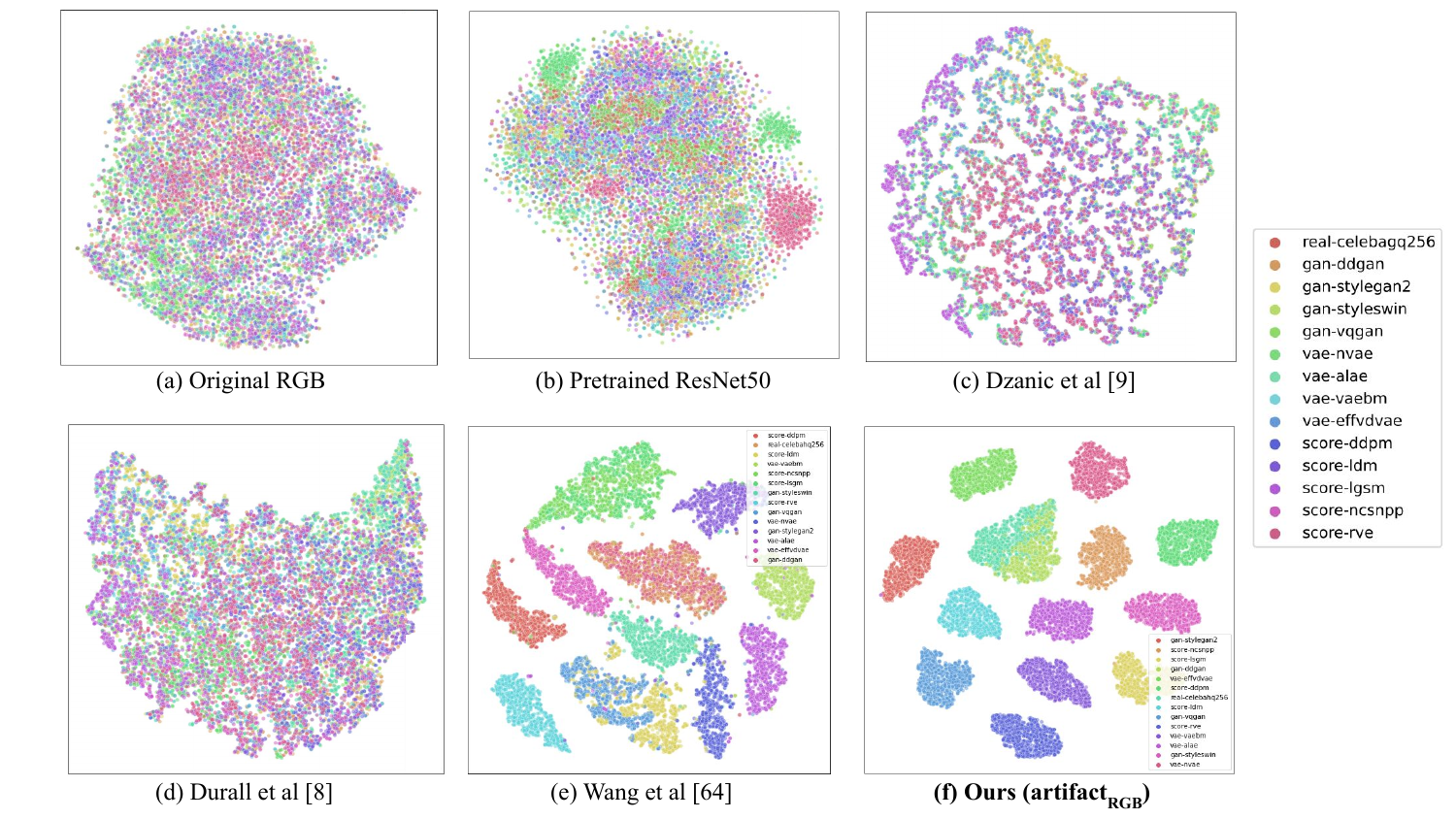} 

    \caption{Features learnt using our definition of artifacts (f) achieve better separation between samples from different generative models (shown in \textit{different colors}). (a) Shows tSNE of generated samples in pixel space, (b) in the latent space of ResNet50 pretrained on ResNet50, (c-f) in the penultimate layer of the classifier proposed by each method trained on the task of model attribution.}
    \label{fig:tsne-gallery}
        \vspace{-1em}

\end{figure*}

%% file: sec/01-intro.tex
While distinguishing synthetic images from real ones has received a lot of attention in recent years  \cite{zhou2018TwoStreamNeuralNetworks, cozzolino2018ForensicTransferWeaklysupervisedDomain, zhu2021FaceForgeryDetection, fung2021DeepfakeUCLDeepfakeDetection,
shen2022IdentityReferencedDeepfakeDetectionb, khormali2022DFDTEndtoEndDeepFake}, distinguishing between different methods of image synthesis remains mostly under-studied~\cite{yu2019AttributingFakeImages, wang2020CNNGeneratedImagesAre}.
The latter, denoted model attribution, is an interesting problem in two respects:
first, in practice it is often valuable to identify the source of synthetic data -- not just that it is synthetic -- to differentiate between authorized vs. malicious personification \cite{giudice2021FightingDeepfakesDetecting} and digital copyright infringement  \cite{franceschelli2022CopyrightGenerativeDeep}; 
second, model attribution provides a systematic way of
studying the similarities and differences between various families of generative models and
revealing the unique limitations of each family in learning the true distribution, 
thereby accelerating developments of new generative models that improve upon such limitations \cite{durall2020WatchYourUpConvolution, wang2020CNNGeneratedImagesAre, chandrasegaran2021CloserLookFourier, guarnera2020DeepFakeDetectionAnalyzing, qian2020ThinkingFrequencyFace}. 

We study the problem of fingerprinting generative models (GMs) based on their samples with an aim to provide a formal, analytical framework to study and compare their characteristics. 
Recent works on GAN-generated images and DeepFake detection have observed
that generative models leave unique traces of computations on their samples that
are distinguishable from natural image generation process (\ie image capturing and processing steps of digital cameras). 
Such observations include checkerboard patterns introduced by deconvolution layers in generator networks \cite{odena2016deconvolution},
spectral discrepancies due to the frequency bias of generative models \cite{schwarz2021FrequencyBiasGenerative, durall2020WatchYourUpConvolution, Marra2019DoGL, wang2020CNNGeneratedImagesAre}
and semantic inconsistencies like mismatched eye colors and asymmetric facial features \cite{mccloskey2018DetectingGANgeneratedImagerya}.


The existence of such fingerprints have been supported by two streams of approaches: one by showing that a classifier can effectively distinguish samples generated by a model from real samples~\cite{wang2020CNNGeneratedImagesAre, yu2019AttributingFakeImages}, and another by directly visualizing the differences between generated and true samples~\cite{dzanic2020FourierSpectrumDiscrepancies, durall2020WatchYourUpConvolution, yu2019AttributingFakeImages}. 
While such works provide evidence for the existence of some fingerprints, the fingerprint itself still remains without an explicit definition. In other words, existing works define the measures of fingerprints~\cite{durall2020WatchYourUpConvolution, dzanic2020FourierSpectrumDiscrepancies, Frank2020LeveragingFA, mccloskey2018DetectingGANgeneratedImagerya, nataraj2019DetectingGANGenerateda},  rather than fingerprints themselves.
This brings about two major drawbacks in developing methods for fingerprinting GMs and studying their characteristics in a principled way. 
First, without a proper definition it is hard to gauge whether the different proposed measures are evaluating the same phenomenon. 
Secondly, and perhaps more importantly, a principled study of the fingerprints themselves -- beyond showing their mere existence -- is challenging.

To this end, the aim of this work is to (i) give a proper definition of generative models' fingerprints, 
(ii) show its sufficiency to capture the notion of fingerprints measured by existing works, 
and finally (iii) study properties of the fingerprints across a diverse set of GMs.  
We find that our proposed definition provides a useful feature space for differentiating generative models among a large array of state-of-the-art (SoTA) models, outperforms existing methods on model attribution (\ie given an image, identify which generative model has generated it), and reveals interesting relations between the models.

By providing a formal definition of GM fingerprints, we address another important gap in literature, which is the lack of studies on fingerprints that are suitable for identifying different generative models in a multi-class setting:
most existing methods including DeepFake detection (\cite{yu2019AttributingFakeImages, Marra2019DoGL, rossler2019faceforensics++, nguyen2022deep}) focus on a binary classification of real vs. synthetic samples.
Thus, whether the existing fingerprints are effective for discriminating \textbf{amongst} various generative methods has not been properly explored. 
To address this limitation, 
we introduce four new benchmark datasets (\gmcifar{}, \gmceleba{}, \gmchq{}, \gmffhq{}),
each constructed from generative models trained on different training datasets (CIFAR-10~\cite{Krizhevsky2009LearningML}, CelebA-64~\cite{liu2015faceattributes}, CelebA-HQ~\cite{Karras2018ProgressiveGO} 
and FFHQ~\cite{Karras2019ASG}, respectively),
and run extensive evaluations on them. 
Our contributions can be summarized as following:
\begin{itemize}
    \item We formalize the definition of fingerprints in generative models, and propose a practical algorithm to compute them from finite samples. 
    \item We provide theoretical justification of our definition by relating it to two prominent metrics for distinguishing generative models: Precision and Recall (P\&R)~\cite{sajjadi2018assessing, kynkaanniemi2019improved} and integral probability metrics (IPMs)~\cite{muller1997IntegralProbabilityMetrics, sriperumbudur2009IntegralProbabilityMetricsa}.
    \item We conduct  extensive experiments to show the effectiveness of our fingerprints in distinguishing generative models, outperforming existing attribution methods. 
    In particular, our experiments consider a large array of generative models from all four main families (GAN, VAE, Flow, Score-based) 
    as opposed to a small number of GANs or VAEs in the exiting works. 
    \item We use the proposed definitions to study the effects of design factors (\eg type of layers) on the  model fingerprints, and observe that the choice of loss functions and upsampling  have the most significant effect. Our findings confirm the general intuition in the research community about the sources of limitations in generative models~\cite{durall2020WatchYourUpConvolution, dzanic2020FourierSpectrumDiscrepancies}, thereby showing the utility of our definitions.
\end{itemize}

%% file: sec/02-rel-work.tex
\textbf{Generative models and their fingerprints.}
Despite the advancement in generative models, recent works on their artifacts and biases have shown that the samples they generate contain features that can be used to identify the source models. 
The existing observations of such features are categorized into four kinds: 
steganalysis-based, color-based, frequency-based and learning-based features. 

\noindent\textbf{Steganalysis-based features.}
Inspired by the methods of fingerprinting digital cameras based on the PRNU patterns (residual patterns), Marra et al.~\cite{Marra2019DoGL} investigates whether GAN image synthesis models leave unique and stable marks on the images they generate and whether they can be used to address image attribution task. 
They propose the residual image as a representation of the image-level fingerprint, and estimates the model-level fingerprint as the average of the residual images by the source generative model. 
We share the same goal as their work in that we want to learn unique, discriminative representations of generative models based on their samples (\ie fingerprinting GMs), 
yet expand the set of GMs under consideration to more diverse and more recent models such as the state-of-the-art GANs (DDGAN) and score-based models like DDPM, NCSN++ and LSGM. 

\noindent\textbf{Color-based features.} 
McCloskey et al.~\cite{mccloskey2018DetectingGANgeneratedImagerya} shows that the histogram of saturated and under-exposed pixels in GAN images provides a sufficient signal to discriminate StyleGAN-generated images from the real images.
Nataraj et al.~\cite{nataraj2019DetectingGANGenerateda} and Nowroozi et al.~\cite{nowroozi2022DetectingHighQualityGANGeneratedb} use co-occurrence matrix on color-bands as the input representation of image fingerprints and feed them to CNN to predict Real vs. GAN images. 

\noindent{}\textbf{Frequency-based features. }
Durall et al.~\cite{durall2020WatchYourUpConvolution}
shows that current generative models  fail to correctly reproduce the spectral distributions of the images in their training dataset by analyzing the spectral statistics of real and synthetic images. 
They propose a hand-crafted, 1-dimensional feature as frequency-based fingerprint feature, computed via azimuthal integration over the 2D spectrum of each image.
They show that this frequency feature is sufficient to distinguish real and synthetic images on the dataset containing AutoEncoders, deepfake manipulation methods and four GANs trained on CelebA (DCGAN, DRAGAN, LSGAN, WGAN-GP).
Dzanic et al.~\cite{dzanic2020FourierSpectrumDiscrepancies} shows that synthetic images manifest different spectral statistics at high-frequencies in two aspects -- the amount of high-frequency contents and the decay rate -- and proposes two parameters that capture these two aspects from a reduced spectru as fingerprints. 
They consider three GANs (StyleGAN1, StyleGAN2, ProGAN) and two VAEs (VQ-VAE2, ALAE). 
However, 
their work is limited to the binary classification 
and do not address model attribution among different generative models.
Wang et al.~\cite{wang2020CNNGeneratedImagesAre}
hypothesizes that CNN-based generators leave common fingerprints 
on the images they generates which can be used to distinguish them from natural images.
This feature can be considered an image-level fingerprint, \ie features on individual images, yet their work does not make an explicit definition of model-level fingerprints.
Our work considers both image-level and model-level fingerprints, 
and further clarifies the two notions by formally defining them in Sec.~\ref{sec:our-defns} as ``GM artifacts'' (for  image-level features) and ``GM fingerprints'' (for model-level features).

\noindent{}\textbf{Learning-based features. }
Marra et al.~\cite{marra2018DetectionGANGeneratedFake}
uses CNN classifier (pretrained and then fine-tuned on their dataset) to predict the real vs. fake images.
Yu et al.~\cite{yu2019AttributingFakeImages} is one of the first works to differentiate the notion of ``image fingerprints'' and ``model fingerprints''.
In their work, ``image fingerprints'' refers to features that exist across sample by the same model, and ``model fingerprints'' refers to features that distinguish one source model from another (e.g., fingerprint of progan-seed0 vs. sngan-seed10).
However, the formal definition of such fingerprints were not provided in their work.

%% file: tbls/fig-our-attribution-workflow.tex
\begin{figure*}[t!]
  \centering
  \includegraphics[width=0.9\linewidth]{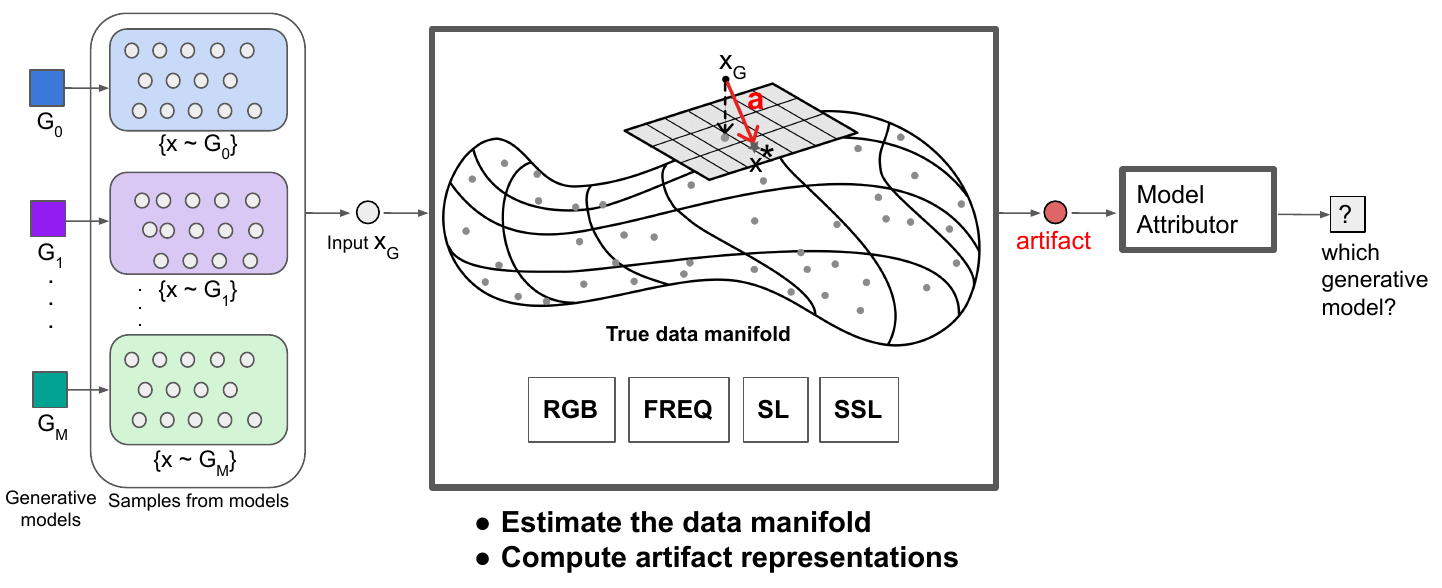}

  \caption{\textbf{Our attribution method.} 
  We propose a model attribution method based on our definition of artifact as deviations from an estimate data manifold. 
  Given input images $X_G$, we first map the images to a chosen embedding space (RGB, Frequency, a feature space of a pretrained supervised-learning (SL) or self-supervised leanring (SSL) network) and compute their artifacts $a$.
  We then pass the artifacts to a ResNet50-based attribution network (Model Attributor) and fine-tune the network to identify the source generative model under the (multi-class) cross-entropy loss.
  }
  \label{fig:our-attribution-network} 

   \vspace{-1em}

\end{figure*}

%% file: sec/03-our-method.tex
As discussed in Sec.~\ref{sec:intro} and Sec.~\ref{sec:rel-work}, despite existing works that provide evidence for the existence of fingerprints and artifacts in generative models, the concrete definitions of these terms remain unclear. In this section, we motivate and propose formal definitions of artifacts and fingerprints of generative models, and then describe a practical algorithm for computing them from observed samples.

\subsection{Definitions of GM artifacts and fingerprints}  \label{sec:our-defns}
Intuitively, the artifacts and fingerprints of generative models are the traces of their imperfection in modeling the generative process of real data, left consistently on the samples they generate. 
From the point of view of manifold learning~\cite{cayton2005algorithms, fefferman2016testing}, which hypothesize that many high-dimensional real-world datasets (\eg images and videos) lie on a lower-dimensional manifold, such \textit{imperfections} of generative models can be formalized in terms of the \textit{deviation of generated samples from the true data manifold}. 
More concretely, consider a generative model $G$ which is trained on a dataset $X$ of real samples that lie on a data manifold $\Phigt$,
with $P_G$ the induced probability distribution of $G$, $S_G$ its support, and $x_G$ a sample generated by $G$:
\begin{definition}[Artifact] \label{defn:artifact}
An artifact $a_{\mathcal{M}}$ left by generative model $G$ on a sample $x_G$ generated by $G$ is defined as the difference between $x_G$ and closest point $x^*$ on the manifold $\mathcal{M}$ of the dataset used to train $G$, 
equipped with a distance metric $d_M$:
\begin{align}
    x^* &:= \argmin_{x \in M}  d_M(x_G, x) \\
    a_{\mathcal{M}}(x_G) &:= x_G - x^*
\end{align}
\end{definition}

\begin{definition}[Fingerprint] \label{defn:fpt}
The fingerprint of a generative model $G$, whose support is $S_G$, with respect to the data manifold $\mathcal{M}$ is defined as the set of all its artifacts:
\begin{align}
    F_G = \{ a_{\mathcal{M}}(x) | x \in S_G \}
\end{align}
\end{definition}
\noindent{}Figure~\ref{fig:our-defn-fpts} illustrates our proposed definitions.

\subsection{Estimation of GM artifacts and fingerprints} \label{subsec:our-defns:how-to-compute}

\noindent\textbf{Estimating the data manifold.}
Since we do not have an access to the data manifold $\mathcal{M}$  on which the real image lie (\ie the natural image manifold), 
we need to estimate it using the observed samples at hand:
To this end, we use real images in the training datasets of the generative models,
and map them to a suitable embedding space to construct a collection of features to be used as an estimated image manifold. 
One key modeling decision in this step is the choice of an embedding space for the image manifold.
We experiment with four possibilities (RGB, Frequency, and learned spaces of a supervised-learning method and a self-supervised learning method) in Sec.~\ref{exp:setup:choice-of-an-embedding-space}.

\vspace{0.5em}
\noindent\textbf{Computing the artifact of a model-generated sample.}
An artifact of a sample $x_G$ is computed in two steps:
\begin{enumerate}
    \item
    Estimate the projection $x^{\star}$ by minimizing the distance to $x_G$ over the points in $X$. We use the Euclidean distance, 
    \ie $d_M(x, x_G) := || x - x_G ||^2$
    \item Compute the artifact as difference, $a(x_G) = x_G - x^*$
\end{enumerate}
Fig.~\ref{fig:viz-our-art-defn} shows some examples of the projections and artifacts in RGB, FREQ, SL and SSL spaces computed in this way.
\vspace{0.5em}

\noindent\textbf{Fingerprint of a model} Given a set of model generated samples $X_G = \{x_i\}_{i=1}^N$ where $x_i \sim P_G$,  we estimate its fingerprint by computing an artifact of each sample in $X_G$, \ie
$F_G = \{ a(x) | x \in X_G \}$.

\subsection{Theoretical justification of our definitions}
Our proposed definitions of GM artifacts and fingerprints are closely related to two prominent metrics for distinguishing generative models: Precision and Recall (P\&R)~\cite{sajjadi2018assessing, kynkaanniemi2019improved} and integral probability metrics (IPMs)~\cite{muller1997IntegralProbabilityMetrics, sriperumbudur2009IntegralProbabilityMetricsa}. 
The most fundamental relation is that under our definition, 
the fingerprint is a non-zero set if and only if two distributions have unequal supports. From this fact several properties of fingerprints under our definition readily follow:

\noindent
\textbf{(A) Relation to Precision and Recall (P\&R)}

\noindent 
Let $P$ denote the true data distribution, $Q$ the generator $G$'s distribution, and $\text{FPT}(Q,P)$ the fingerprint of $G$ w.r.t. $P$ as defined in \ref{sec:our-defns}.
Let $d_\text{FPT}(Q, P)$ be the norm of a largest artifact vector in $\text{FPT}(Q,P)$ defined as,
\begin{equation} 
  d_\text{FPT}(Q, P) := \sup_{x_G \sim Q} \{  ||a||_2 : a \in FPT(Q,P )\} \label{defn:d_fpt}
\end{equation}
\noindent $d_\text{FPT}(Q, P)$ is one way to quantify the maximal deviation of the generator's manifold (i.e., $\text{Supp}(Q)$) from the data manifold (i.e., $\text{Supp}(P)$). Note that  $d_\text{FPT}(Q, P) \geq 0$.
First of all, the following equivalences hold:
\begin{align}
  &\text{"All images $x_G$ from $G$ lie on the true data manifold"} \notag \\
   &\Leftrightarrow \forall x_G \sim Q:  x_G \in \text{Supp}(P) \Leftrightarrow x^{\star} = x_G \\ 
   &\Leftrightarrow \forall x_G \sim Q:  a(x_G) = \vec{0} ~~~~\text{(by Eqn.2)}  \\ 
  &\Leftrightarrow \text{FPT}(Q,P) = \{ \vec{0}\}  \\
  &\Leftrightarrow d_\text{FPT}(Q, P) = 0     
\end{align}

\noindent By the definition of P\&R in Defn (2) of \cite{kynkaanniemi2019improved}, 
\begin{equation}
  \forall x_G \sim Q:  x_G \in \text{Supp}(P) \Leftrightarrow \text{Precision}(Q,P) = 1     \label{prop:max-precision}
\end{equation}

\noindent Therefore, $d_\text{FPT}(Q, P) = 0 \Leftrightarrow \text{Precision}(Q,P) = 1$, and the minimum achievable deviation of $Q$ from $P$ based on our definitions of artifacts and fingerprints corresponds to the maximal achievable precision.

Similarly, by considering $\text{FPT}$ of $P$ with respect to $Q$ where $Q$ is now the reference distribution,
\begin{align}
  \text{FPT}(P,Q) = \{ \vec{0}\}  &\Leftrightarrow d_\text{FPT}(P,Q) = 0  \\
                                  &\Leftrightarrow \text{Recall}(Q,P) = 1
\end{align}
\noindent In other words, the minimum achievable deviation of $P$ from $Q$ corresponds to the maximal recall.

In summary, the following relationships between our definition of fingerprint and P\&R hold:
\begin{align}
  &\text{FPT}(Q,P) = \{ \vec{0}\} \Leftrightarrow \text{Precision}(Q,P) = 1  ~~\textbf{(max precision)}   \label{prop:max-precision}\\
    &\text{FPT}(P,Q) = \{ \vec{0}\}   \Leftrightarrow ~~\text{Recall}(Q,P) = 1 ~~\textbf{(max recall)}  \label{prop:max-recall}
\end{align}
\noindent Additionally, we have the property of equal supports:
\begin{align}
    &\text{FPT}(Q,P) = \{ \vec{0}\} ~~\text{and} ~~\text{FPT}(P,Q) = \{ \vec{0}\}& \notag \\
  \Leftrightarrow &  ~~~~~~~~~~~~\text{Supp(P)} = \text{Supp}(Q)    ~~~~\textbf{(equal supports)} \label{prop:equal-supports} 
\end{align}
\noindent The property of equal supports implies the degree of our fingerprint's expressivity in regards to the difference between $P$ and $Q$: as long as there is at least one generated sample not on the data manifold, our fingerprint (either Q w.r.t P or P w.r.t Q) can encode that difference by having at least one non-zero element and its $d_{\text{FPT}}$ strictly greater than zero.

\noindent
\textbf{(B) Relation to integral probability metrics (IPMs)}
Our definition of fingerprint is related to integral probability metrics (IPMs)~\cite{muller1997IntegralProbabilityMetrics, sriperumbudur2009IntegralProbabilityMetricsa}, which include MMD and Wasserstein distance, in the following way:
By the property of equal supports in Eqn.~\ref{prop:equal-supports},
\begin{equation}
  \text{Supp}(P) \neq \text{Supp}(Q) \Leftrightarrow \exists a \in \text{FPT}(Q,P) \neq \vec{0}  \label{eqn:a} 
\end{equation}

\noindent By the definition of IPMs \cite{muller1997IntegralProbabilityMetrics, sriperumbudur2009IntegralProbabilityMetricsa},
\begin{equation}
  \text{Supp}(P) \neq \text{Supp}(Q) \Rightarrow \exists \text{IPM}(Q,P) \neq 0  \label{eqn:b}  
\end{equation}

\noindent From Eqn.\ref{eqn:a} and Eqn.\ref{eqn:b}, we have:
\begin{align}
  &\exists a \in \text{FPT}(Q,P) \neq \vec{0}  ~~\text{(\ie $d_{\text{FPT}} \neq 0)$}  \notag \\  
  &\Rightarrow \exists \text{IPM}(Q,P) \neq 0  \label{eqn:c} 
\end{align}

\noindent Conversely,
\begin{align}
  \forall \text{IPM}: \text{IPM}(Q,P) = 0 \Rightarrow  \text{FPT}(Q,P) = \{ \vec{0} \} \notag \\
  \text{(\ie $d_{\text{FPT}}(Q,P)= 0$)} \label{eqn:d}  
\end{align}

\noindent This means if all IPMs vanish to zero, our fingerprint also vanishes to a trivial set that only contains a zero vector.

\subsection{Attribution network}
The goal of our attribution network is to predict the source generative model of an observed image.
It takes as input our artifact representation of an image (as defined in \ref{sec:our-defns}) and predicts the identity of its source generative model. 
We use ResNet50~\cite{he2016deep} as our attribution network and finetune it with the standard cross-entropy loss.
The network is trained for source model attribution over images generated by different generative models. 
The first step in our attribution procedure is to represent the input image as an artifact feature by computing its deviation from an estimate data manifold as discussed in Sec.~\ref{subsec:our-defns:how-to-compute}.
Note that the main novelty in our attribution method is in representing each input as an artifact, \ie a deviation from the estimated data manifold, where we consider either RGB, frequency, or ResNet-learned feature space as the embedding space of the manifold (Sec~\ref{subsec:our-defns:how-to-compute}). 
Figure~\ref{fig:our-attribution-network} illustrates our attribution process.

%% file: tbls/fig-viz-our-defns.tex
\begin{figure*}[ht!]
\centering
{\scriptsize \def\teaserwid{0.14\linewidth}  

\begin{tabular}{c@{\hspace{.5mm}}*{7}{c@{\hspace{.5mm}}}}

    \rotatebox[origin=c]{90}{DDGAN~\cite{xiao2022TACKLINGGENERATIVELEARNING}}&
     \raisebox{-0.45\height}{\includegraphics[width=\teaserwid]{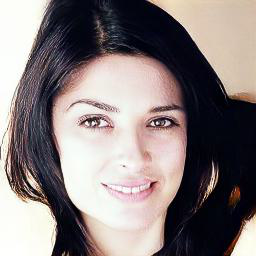}}&  
     \raisebox{-0.45\height}{\includegraphics[width=\teaserwid]{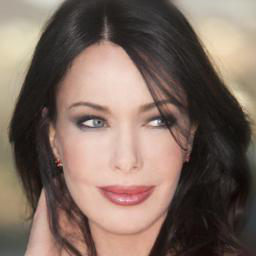}}&
     \raisebox{-0.45\height}{\includegraphics[width=\teaserwid]{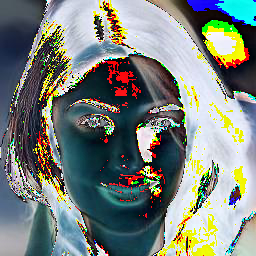}}&
     \raisebox{-0.45\height}{\includegraphics[width=\teaserwid]{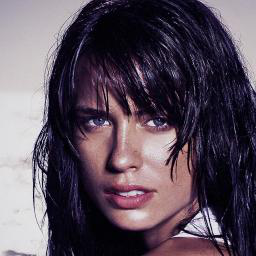}}&
     \raisebox{-0.45\height}{\includegraphics[width=\teaserwid]{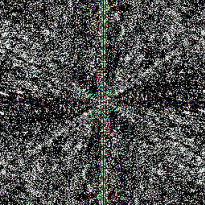}}&
     \raisebox{-0.45\height}{\includegraphics[width=\teaserwid]{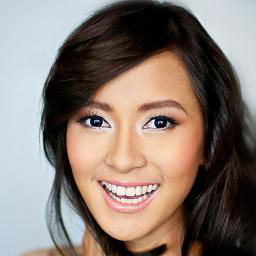}}&
     \raisebox{-0.45\height}{\includegraphics[width=\teaserwid,height=\teaserwid]{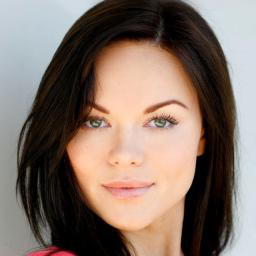}}\\[1cm]  
     
    \rotatebox[origin=c]{90}{Eff-VDVAE~\cite{hazami2022EfficientVDVAELessMore}}&
     \raisebox{-0.45\height}{\includegraphics[width=\teaserwid]{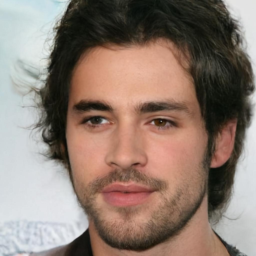}}&  
     \raisebox{-0.45\height}{\includegraphics[width=\teaserwid]{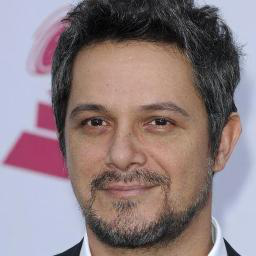}}&
     \raisebox{-0.45\height}{\includegraphics[width=\teaserwid]{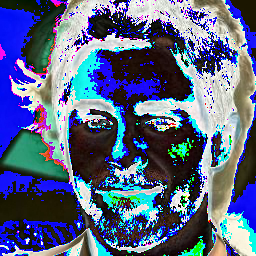}}&
     \raisebox{-0.45\height}{\includegraphics[width=\teaserwid]{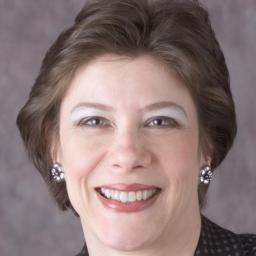}}&
     \raisebox{-0.45\height}{\includegraphics[width=\teaserwid]{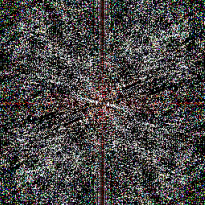}}&
     \raisebox{-0.45\height}{\includegraphics[width=\teaserwid]{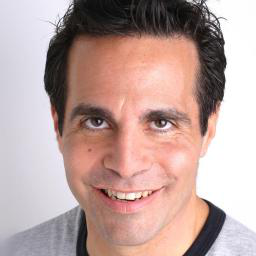}}&
     \raisebox{-0.45\height}{\includegraphics[width=\teaserwid,height=\teaserwid]{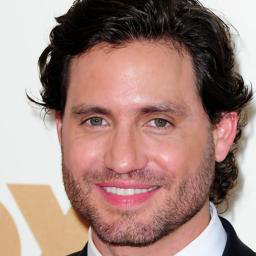}}\\[1cm] 

    \rotatebox[origin=c]{90}{NCSN++~\cite{song2023ScoreBasedGenerativeModeling}}&
     \raisebox{-0.45\height}{\includegraphics[width=\teaserwid]{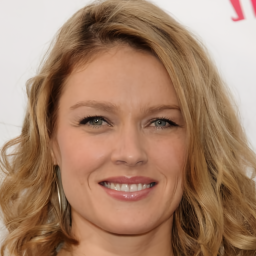}}&  
     \raisebox{-0.45\height}{\includegraphics[width=\teaserwid]{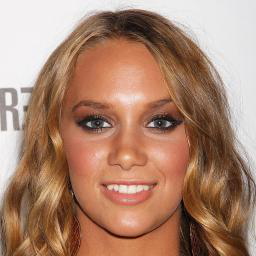}}&
     \raisebox{-0.45\height}{\includegraphics[width=\teaserwid]{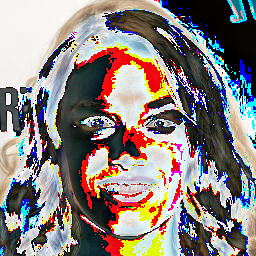}}&
     \raisebox{-0.45\height}{\includegraphics[width=\teaserwid]{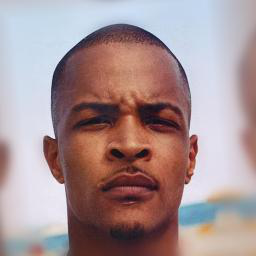}}&
     \raisebox{-0.45\height}{\includegraphics[width=\teaserwid]{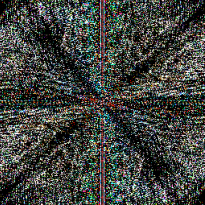}}&
     \raisebox{-0.45\height}{\includegraphics[width=\teaserwid]{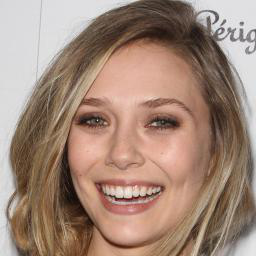}}&
     \raisebox{-0.45\height}{\includegraphics[width=\teaserwid,height=\teaserwid]{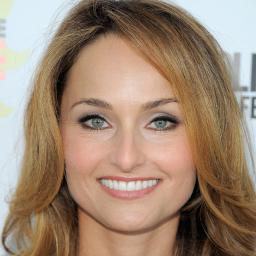}}
    \vspace{.05in}
     \\
     &
      Generated image  &  
     $\phirgb^{\star}$  &  
      artifact in RGB &  
     $\phifreq^{\star}$  &  
     artifact in FREQ &  
     $\phisl^{\star}$  &  
     $\phissl^{\star}$  
\end{tabular}
}
\caption {
We visualize artifacts in generated images under our manifold-based definition (Sec.~\ref{defn:artifact}). 
Each row shows an original image generated by a generative model, followed by its projection to data manifolds in RGB ($\phirgb^{\star}$), Frequency ($\phifreq^{\star}$), and learned feature spaces of SL ($\phisl^{\star}$) and SSL ($\phissl^{\star}$).  
The third and fourth columns show our definition of artifacts in the RGB and frequency spaces, respectively. 
Note that artifacts in SL and SSL spaces are not shown as they are 2048-long vectors (in the embedding space of a pretrained ResNet50).
}
\label{fig:viz-our-art-defn}
 \vspace{-1em}
\end{figure*}

%% file: sec/04-exps.tex
\noindent{}We start our experiments by hypothesizing the existence of fingerprints for individual generative models.
Sec.~\ref{subsec:datasets-baselines} explains our experimental setup. 
Sec.~\ref{subsec:exp:classify} and Sec.~\ref{subsec:exp:feature-space-analysis} explore the existence of GM fingerprints and the attributability of generative models via classification and feature space analysis.
Building on experimental supports for their existence,
Sec.~\ref{subsec:exp:characterize}  studies the clustering structure of the set of GM artifacts, and their relation to different generative models.

\subsection{Experimental setup} \label{subsec:datasets-baselines}
\input{tbls/tbl-exp-all-gmsets}

\input{rqs/exp-datasets}

\input{rqs/exp-baselines} %

\input{tbls/tbl-exp-result-1-multiclass}  

\input{rqs/exp-setup-choice-of-embedding-spaces}
\input{tbls/tbl-our-notation} 

\subsection{Existence of fingerprints in generative models} \label{subsec:exp:classify}   
\input{rqs/rq1-existence}

\subsection{Feature space analysis} \label{subsec:exp:feature-space-analysis}
\input{rqs/rq1c-feature-space-analysis}

\input{tbls/tbl-exp-result-3-a-generalize} 

\subsection{Cross-dataset generalization}
 \label{subsec:exp:cross-dataset-generalize}
\input{rqs/rq3-generalize}

\input{tbls/tbl-exp-result-2-layer-types} 

\subsection{Clustering structure of GM fingerprints}
\label{subsec:exp:characterize} 
\input{rqs/rq2-clustering-structure-of-artifacts}


%% file: tbls/tbl-exp-all-gmsets.tex
\begin{table}[t]
\centering 
\resizebox{1.\linewidth}{!}{
{\def\arraystretch{1.3}
\begin{tabular}{@{}l cccc @{}}
\toprule 
    \textbf{Family} & \textbf{\gmcifar{}} & \textbf{\gmceleba{}} & \textbf{\gmchq{}}  & \textbf{\gmffhq{}} \\
\midrule
    \textbf{Real} &CIFAR-10~\cite{Krizhevsky2009LearningML}  &CelebA~\cite{liu2015faceattributes} &CelebA-HQ (256)~\cite{Karras2018ProgressiveGO} &FFHQ (256)~\cite{Karras2019ASG} \\   
\hdashline  
    \textbf{GAN}  
    & BigGAN-Deep~\cite{brock2023LargeScaleGAN}  & plain GAN~\cite{Goodfellow2014GenerativeAN} & BigGAN-Deep~\cite{brock2023LargeScaleGAN} & BigGAN-Deep~\cite{brock2023LargeScaleGAN} \\
    & StyleGAN2~\cite{karras2020AnalyzingImprovingImagea}  & DCGAN~\cite{Radford2016UnsupervisedRL} & StyleGAN2~\cite{karras2020AnalyzingImprovingImagea} & StyleGAN2~\cite{karras2020AnalyzingImprovingImagea}  \\
    & & LSGAN~\cite{Mao2017LeastSG} & StyleGAN3~\cite{Karras2021AliasFreeGA} & StyleGAN3~\cite{Karras2021AliasFreeGA} \\ 
    & WGAN-gp~\cite{Gulrajani2017ImprovedTO}  & WGAN-gp/lp~\cite{Gulrajani2017ImprovedTO} & VQ-GAN~\cite{esser2021TamingTransformersHighResolutiona} & VQ-GAN~\cite{esser2021TamingTransformersHighResolutiona} \\ 
    & & DRAGAN-gp/lp~\cite{kodali2017convergence} & StyleSwin~\cite{zhang2022StyleSwinTransformerbasedGANa} & \\ 
    & DDGAN~\cite{xiao2022TACKLINGGENERATIVELEARNING}  &  & DDGAN~\cite{xiao2022TACKLINGGENERATIVELEARNING} & \\ 
\hdashline  
    \textbf{VAE}         
    & & $\beta$-VAE~\cite{Higgins2017betaVAELB} & & \\
    & & DFC-VAE~\cite{Hou2017DeepFC} & StyleALAE~\cite{pidhorskyi2020AdversarialLatentAutoencoders} &   \\ 
    & NVAE~\cite{vahdat2020NVAEDeepHierarchical}  & NVAE~\cite{vahdat2020NVAEDeepHierarchical} & NVAE~\cite{vahdat2020NVAEDeepHierarchical} & NVAE~\cite{vahdat2020NVAEDeepHierarchical} \\  
    & VAE-BM~\cite{xiao2022VAEBMSymbiosisVariational}  & VAE-BM~\cite{xiao2022VAEBMSymbiosisVariational} & VAE-BM~\cite{xiao2022VAEBMSymbiosisVariational} &  \\  
    & Eff-VDVAE~\cite{hazami2022EfficientVDVAELessMore}  & Eff-VDVAE~\cite{hazami2022EfficientVDVAELessMore} & Eff-VDVAE~\cite{hazami2022EfficientVDVAELessMore} & Eff-VDVAE~\cite{hazami2022EfficientVDVAELessMore} \\  
\hdashline  
    \textbf{Flow}
    & GLOW~\cite{kingma2018GlowGenerativeFlow}  & GLOW~\cite{kingma2018GlowGenerativeFlow} &  &   \\
    & MaCow~\cite{Ma2019MaCowMC}  &  & MaCow~\cite{Ma2019MaCowMC} &   \\
    & &  & Residual Flow~\cite{chen2019ResidualFlowsInvertible} &   \\  
\hdashline  
    \textbf{Score}            
    & DDPM~\cite{ho2020DenoisingDiffusionProbabilistic}  & DDPM~\cite{ho2020DenoisingDiffusionProbabilistic} & DDPM~\cite{ho2020DenoisingDiffusionProbabilistic}       &      \\
    & &  & NCSN++~\cite{song2023ScoreBasedGenerativeModeling}      &  NCSN++~\cite{song2023ScoreBasedGenerativeModeling} \\ 
    & RVE~\cite{kim2022SoftTruncationUniversal}  & RVE~\cite{kim2022SoftTruncationUniversal}  & RVE~\cite{kim2022SoftTruncationUniversal}  &  \\  
    & LSGM~\cite{vahdat2021ScorebasedGenerativeModeling}  &  & LSGM~\cite{vahdat2021ScorebasedGenerativeModeling}   &  \\  
    & &  & LDM~\cite{rombach2022HighResolutionImageSynthesisb}  & LDM~\cite{rombach2022HighResolutionImageSynthesisb} \\  
\bottomrule
\end{tabular}}
}
\caption{{\bf Our experimental dataset of generation models.}
We collect images from a diverse set of generative models trained on four different datasets (CIFAR10, CelebA, CelebA-HQ(256), FFHQ(256)) and study model fingerprints and their attributability.
\textbf{Real}: training datasets of the generative models. 
\textbf{Score}: score-based (aka. diffusion) models. 
}
\label{tbl:exp-datasets}
\vspace{-5mm}
\end{table}

%% file: rqs/exp-datasets.tex
\noindent\textbf{Datasets} \label{sec:exp:datasets}
To evaluate how well different fingerprints perform on model attribution in a truly multi-class setting, we propose four new datasets -- \gmcifar{}, \gmceleba{}, \gmchq{} and \gmffhq{}  -- constructed from real datasets and generative models trained on CIFAR-10~\cite{Krizhevsky2009LearningML}, CelebA-64~\cite{liu2015faceattributes}, CelebA-HQ(256)~\cite{Karras2018ProgressiveGO} 
 and FFHQ(256)~\cite{Karras2019ASG}, respectively. 
We collect 100k images from each generative model.
Our datasets address the absence of benchmark datasets for studying the attribution of generative models
by including a variety of models from GAN, VAE, Flow and Score-based family,
and state-of-the-art generative models (\eg DDGAN, VAE-BM, LSGM) that have not been considered before.
Tab.~\ref{tbl:exp-datasets} summarizes our datasets, organized in column by the training datasets.
See Appendix~\ref{appendix:our-datasets} for details on our dataset creation process.
 We highlight that each dataset exclusively consists of models trained on the same training dataset: 
this consistency is important for studying the effects of model architectures and datasets on attribution independently, which we study in Sec.~\ref{subsec:exp:classify} and Sec.~\ref{subsec:exp:cross-dataset-generalize}.

%% file: rqs/exp-baselines.tex
\noindent\textbf{Baselines} \label{sec:exp:baselines}
Existing methods of fingerprinting generative models can be categorized into three groups:
color-based, frequency-based and supervised-learning.
We consider key methods from each group
and compare them to our proposed method.
See details on the baselines in Appendix~\ref{appendix:baselines}. 
\begin{itemize}
    \item Color-based: 
Histogram of saturated, under-exposed pixels \cite{mccloskey2018DetectingGANgeneratedImagerya}, 
Co-occurrence matrix \cite{nataraj2019DetectingGANGenerateda}
    \item Frequency-based:
1-dim power spectrum via azimuthal integration on DCT \cite{durall2020WatchYourUpConvolution},
high-frequency decay parameters fitted to normalized reduced spectra \cite{dzanic2020FourierSpectrumDiscrepancies}
    \item Features via supervised-learning:
InceptionNet-v3~\cite{marra2018DetectionGANGeneratedFake}, XceptionNet~\cite{marra2018DetectionGANGeneratedFake},
Yu19~\cite{yu2019AttributingFakeImages},
Wang20~\cite{wang2020CNNGeneratedImagesAre}

\end{itemize}

%% file: tbls/tbl-exp-result-1-multiclass.tex
\begin{table*}[t]
\setlength{\tabcolsep}{.5em}

\centering
\small
\resizebox{0.95\linewidth}{!}{
      \begin{tabular}{@{}lc cc cc cc cc@{}}
\toprule
    && \gmcifar{} & & \gmceleba{}  &  & \gmchq{} &  & \gmffhq{}  \\
	\cline{3-10}\vspace{-1em}\\
    Methods && Acc.(\%)$\uparrow$  & FDR$\uparrow$ & Acc.(\%)$\uparrow$  & FDR$\uparrow$ & Acc.(\%)$\uparrow$  & FDR$\uparrow$  &  Acc.(\%)$\uparrow$  & FDR$\uparrow$ \\
\toprule
    McClo18~\cite{mccloskey2018DetectingGANgeneratedImagerya}  && $40.223\pm1.10$ & 32.4  & $62.6 \pm 2.314$ & 70.2 & $57.4 \pm 2.013$ & 36.3  & $50.8 \pm 0.341$ & 26.3\\
    Nataraj19~\cite{nataraj2019DetectingGANGenerateda}  && $46.291\pm1.43$  & 36.7 & $61.1 \pm 2.203$ & 74.0 & $56.3 \pm 1.325$ &37.9 &$51.3 \pm 0.581$ &35.3\\
    Durall20~\cite{durall2020WatchYourUpConvolution}  && $57.293\pm0.93$  & 46.5 & $62.2 \pm 2.243$ & 75.5 & $59.1 \pm 1.301$ &38.8 &$60.9 \pm 0.255$ &37.9 \\
    Dzanic20~\cite{dzanic2020FourierSpectrumDiscrepancies} && $56.123\pm1.21$ & 43.1  & $61.6 \pm 2.029$ & 88.1 & $56.9 \pm 1.215$ &38.2 &$55.7 \pm 0.324$ &30.3 \\
\hline
    Wang20~\cite{wang2020CNNGeneratedImagesAre}  && $62.23\pm0.84$ & 53.6 & $62.2 \pm 1.203$ & 89.8 & $59.5 \pm 1.252$ &30.3  &$64.2 \pm 0.310 $ &37.9\\
    Marra18~\cite{marra2018DetectionGANGeneratedFake}  && $55.944\pm1.09$ & 41.2 & $63.1 \pm 1.103$ & 83.4 & $51.3 \pm 1.281$  &20.5  &$53.2 \pm 0.218$ &30.4\\
\hline 
    Marra19~\cite{marra2019IncrementalLearningDetectiona}  && $60.71\pm1.24$  & 47.2 & $61.1 \pm 1.729$ & 101.4 & $59.1 \pm 1.27$ & 34.9 &$51.8 \pm 0.233$ &30.9 \\
    Yu19~\cite{yu2019AttributingFakeImages}  && $62.01\pm0.79$ & 50.1 & $60.6 \pm 1.103$ & 111.4 & $61.1 \pm 1.122$  &\underline{74.5}  &$60.5 \pm 0.105$ &35.1\\
\hline 
    $\oursrgb{}$ && $69.48\pm1.08$ & 55.2 & $70.5 \pm 1.565$ & 115.3 & $\underline{63.7} \pm 1.238$ & 64.2   &$ \underline{65.3} \pm 0.125$ & \underline{50.1}\\
    $\oursfreq{}$ && $\underline{70.191}\pm0.96$ & \underline{57.2} & $72.8 \pm 1.321$ & 120.9 & $\textbf{64.8} \pm 1.124$ & 70.1 & $\textbf{66.1} \pm 0.207$ & \textbf{57.6} \\
    $\ourssup{}$ && $\textbf{72.018}\pm0.92$ & \textbf{58.9} & $\underline{73.6}\pm 1.102 $ & \textbf{168.0 }  & $62.3 \pm 1.221$ & \textbf{77.2}  &$ 63.2 \pm 0.305$ &49.8\\
    $\oursunsup$ && $70.177\pm1.13$ & 56.1 & $\textbf{74.7} \pm 1.121$ & $\underline{125.9}$ &$ 61.9\pm1.351$ & 63.3  &$ 63.8 \pm 0.203$ &40.9\\
\bottomrule
\end{tabular}
} 
\caption{\small
\textbf{Model attribution results. }
We evaluate different artifact features on the task of predicting the source generative model of a generated sample.
Separability of the feature spaces are measured in FD ratio (FDR). Higher FDR means better separability.
Our methods (\textit{ART}'s) based on the proposed definition of artifacts outperform all baseline methods on four different datasets.
}
\label{tab:exp:multi-class-results}
\end{table*}

%% file: rqs/exp-setup-choice-of-embedding-spaces.tex
\noindent \textbf{Choice of the embedding space} \label{exp:setup:choice-of-an-embedding-space}
One main modeling decision to make when computing our fingerprints (Sec.~\ref{subsec:our-defns:how-to-compute}) is the choice of the embedding space in which the true data manifold (\ie natural image manifold) sit. 
We consider four representation spaces based on the previous works that suggest the existences of fingerprints ~\cite{mccloskey2018DetectingGANgeneratedImagerya, dzanic2020FourierSpectrumDiscrepancies} and visual features~\cite{szegedy2016RethinkingInceptionArchitecture, zbontar2021barlow} encoded in them:
RGB, Frequency, and feature spaces learned by a supervised-learning method (SL) (\eg ResNet50~\cite{he2016deep}) and by a self-supervised learning method (SSL) (\eg Barlow Twins~\cite{zbontar2021barlow}).
To map images to each space, we apply the following transformations (Tab.~\ref{tbl:embedding-maps}).
\begin{itemize}
    \item For RGB space, we use the RGB images as is. 
    \item For frequency space, we transform the RGB images to 2D spectrum by applying the Fast Fourier Transform (FFT) on each channel. 
    \item For the embedding space of a supervised-learning method (SL), we use the encoder head of ResNet50 \cite{he2016deep} pretrained on ImageNet.
    \item For the embedding space of a self-supervised learning method (SSL), we use  the encoder head of the pretrained Barlow Twins~\cite{zbontar2021barlow}.
\end{itemize}

%% file: tbls/tbl-our-notation.tex
\begin{table}[t]
  \centering
  \begin{tabular}{@{}lc@{}}
    \toprule
    Embedding Space & Embedding map \\
    \midrule
    RGB & Identity \\
    FREQ & Channelwise FFT \\
    SL & Pretrained ResNet50 \\
    SSL & Pretrained BarlowTwin \\
    \bottomrule
  \end{tabular}
  \caption{
  \textbf{Our embedding spaces.}
  To estimate the data manifold in a suitable embedding space, we apply each transformation to input images.}
  \label{tbl:embedding-maps}
  \vspace{-1em}
\end{table}

%% file: rqs/rq1-existence.tex
We test the existence of fingerprints on the generated samples by training a classifier for model attribution. 
A high test accuracy implies the classifier is able to extract features from the images that are distinct signatures of their source  models, thereby supporting the existence of their fingerprints. 

\noindent{}\textbf{Metrics. }
We evaluate the attributability of baseline methods listed in Tab.~\ref{tbl:baselines-datasets} and our proposed methods on our GM datasets.
The performance is measured in accuracy(\%).
We consider four variants of our attribution method by representing the artifacts in different embedding  spaces (\{RGB, Frequency, Supervised-learning (SL) and self-supervised learning (SSL)\} spaces; Sec.~\ref{sec:our-defns}).
They are referred to as $\oursrgb$, $\oursfreq$, $\ourssup$, and $\oursunsup$ in Tab.~\ref{tab:exp:multi-class-results}. 

\noindent\textbf{Evaluation Protocol.}
Each dataset consists of real images and generated images from $M$ generative models, and each image is labelled with the identity of their source model, \eg 0 for Real, 1 for $G_1$, ..., M for $G_M$.
We split the data into train, val, test in ratio of $7:2:1$, 
train the classifier on the train split with the cross-entropy loss over the labels, 
and measure the accuracy on the test split. 

\noindent\textbf{Results.}
Tab.~\ref{tab:exp:multi-class-results} shows the result of model attribution using each fingerprinting method. 
First of all, we notice that the overall accuracy is lower on \gmchq, than on \gmceleba . This aligns with our intuition that attributing samples becomes harder when they are generated by more advanced models that match the true data-generating process better. 
Another possible reason for this phenomenon is that SoTA models in \gmchq~ are often hybrid, meaning a model (\eg DDGAN) incorporates architectural and optimization
techniques from multiple families of GMs (\eg a combination of adversarial training and diffusion sampling), thereby making the artifacts the models generate also become a mixture, and harder to attribute to a single model instance.
Secondly, we observe that the fingerprints based on hand-crafted features such as the histogram of saturated,
under-exposed pixels~\cite{mccloskey2018DetectingGANgeneratedImagerya}, co-occurrence matrix of RGB pixels~\cite{nataraj2019DetectingGANGenerateda} and 1D power spectrum~\cite{durall2020WatchYourUpConvolution} perform worse than the fingerprints learned with CNN-based classifiers(\cite{marra2018DetectionGANGeneratedFake}, \cite{yu2019AttributingFakeImages}). 
Lastly, our attribution method outperforms all the existing methods on all datasets by meaningful margins (11.6\%, 3.7\%, 1.9\% in each dataset, and 5.73\% on average), thus supporting our definitions' usefulness as fingerprints of generative models.

%% file: rqs/rq1c-feature-space-analysis.tex
\noindent \textbf{Separability (FD ratio).} 
To complement the accuracy, we measure the separability of fingerprint representations using the ratio of inter-class and intra-class Fréchet Distance (FDR)~\cite{dowson1982frechet}.
The larger the ratio, the more attributable the fingerprints are to their model-type. See Appendix~\ref{appendix:fdr} for the definition of FDR and how to compute it.
Tab.~\ref{tab:exp:multi-class-results} shows the FD ratios computed for the fingerprint representations on the test datasets. 
The FDRs are significantly higher for learned representations (Row of Wang20 and below) than color-based (McClo8, Nataraj19) or frequency-based fingerprints (Durall20, Dzanic20).
In particular, our artifact-based feature spaces achieve improved FDRs, in alignment with the attribution results in classification accuracy.

\noindent \textbf{tSNE of fingerprint features.} We qualitatively compare the feature spaces of six different fingerprint representations using t-SNE~\cite{Maaten2008VisualizingDU} in Fig.~\ref{fig:tsne-gallery}. 
While both the original RGB features or features extracted by a pretrained ResNet50~\cite{he2016deep} show no clear clustering, 
the features learned using our artifact representations (Fig.~\ref{fig:tsne-gallery}.f) show much better separated clusters.
Note that each cluster correctly corresponds to its source generative model, which supports the utility of our features as fingerprints of the generative models.

%% file: tbls/tbl-exp-result-3-a-generalize.tex
\begin{table}[t!]

\centering
\resizebox{\columnwidth}{!}{

\begin{tabular}{lc cc cc}
\toprule
Methods  && C10$\rightarrow$CA   & CA$\rightarrow$C10 & CHQ$\rightarrow$FFHQ   & FFHQ$\rightarrow$CHQ  \\ 
\toprule
    McClo18~\cite{mccloskey2018DetectingGANgeneratedImagerya}  && $52.3$ & $43.2$ & 34.2 & 31.2 \\
    Nataraj19~\cite{nataraj2019DetectingGANGenerateda}  && $56.2$ & $46.1$ & 42.1 &40.4 \\
    Durall20~\cite{durall2020WatchYourUpConvolution}  && $60.1$ & $53.5$ & 51.9 & 42.6 \\
    Dzanic20~\cite{dzanic2020FourierSpectrumDiscrepancies} && $56.9$ & $54.7$ & 45.2 & 42.5 \\
\hline
    Wang20~\cite{wang2020CNNGeneratedImagesAre}  && $62.5$ & $60.1$ & 61.4 & 53.4\\
    Marra18~\cite{marra2018DetectionGANGeneratedFake} && $57.0$  & $58.4$ & 50.2	& 35.9\\
\hline 
    Marra19~\cite{marra2019IncrementalLearningDetectiona}  && $61.0$& $58.6$ & 54.3	& 30.3\\
    Yu19 \cite{yu2019AttributingFakeImages}  && $60.5$ &  $60.4$ & 55.2 &	50.3 \\
\hline 
    $\oursrgb{}$   && $62.7$ &  $60.2$ & \textbf{66.3}	& 53.2 \\
    $\oursfreq{}$  && $65.8$ &  $\underline{62.1}$ & \underline{63.5} &	\underline{54.3}\\
    $\ourssup{}$   && $\textbf{67.7}$   & $60.3$  & 57.6  &	\textbf{56.9}\\
    $\oursunsup{}$ && $\underline{66.3}$  & $\textbf{63.0}$  & 58.1 &	53.5 \\
\bottomrule 
\end{tabular}
}
\caption{
\textbf{Generalization of model attribution across datasets.}
We evaluate how well baselines and our fingerprints generalize across training datasets.  
 We consider two scenarios: (i) generalization across \gmcifar{} and \gmceleba{}, and (ii) generalization across \gmchq{} and \gmffhq{}.  For each case, we train attribution methods on one set of generative models (\eg \gmcifar{}) and test on a different set of models (\eg \gmceleba{}). 
 Our artifact-based attribution method outperforms all baseline methods in both scenarios. 
 \textbf{C10}: CIFAR-10.
 \textbf{CA}: CelebA.
 \textbf{CHQ}: CelebA-HQ.
}
\label{tab:exp:generalize-result-a}
\vspace{-1em}
\end{table}

%% file: rqs/rq3-generalize.tex
We  study the generalizability of  attribution methods across  datasets. 
This generalizability is important in practice, when, \eg a correct attribution is needed even when different end users train new models using their own datasets. 
We consider two scenarios:
generalization (i) across \gmcifar{} and \gmceleba{}, and (ii) across \gmchq{} and \gmffhq{}.

\noindent\textbf{Evaluation.}
In each scenario, we train attribution methods on the training dataset (\eg \gmcifar{}) and test their accuracies on an unseen dataset (\eg \gmceleba{}). 

\noindent\textbf{Results.}
Tab.~\ref{tab:exp:generalize-result-a}  shows the result of the accuracies of the cross-dataset generalization for  \gmcifar{} $\xleftrightarrow{}$ \gmceleba{} and for  \gmchq{} $\xleftrightarrow{} $ \gmffhq{}. 
First of all, our manifold-base attribution methods outperform existing methods in both scenarios.
Note that CIFAR-10 and CelebA contain images from different domains (CIFAR-10: objects and animals vs. CelebA: human faces), while CelebA-HQ and FFHQ both contains facial images. 
Therefore, 
the high accuracies in both scenarios indicate that our  method generalize not only across datasets of similar semantics (CHQ $\xleftrightarrow{}$ FFHQ), but also across of different semantics (CIFAR-10 $\xleftrightarrow{}$ CelebA).
Overall, these higher accuracies show that our methods are more robust to the change of training datasets of the generative models, supporting the efficacy of our methods in practice 
where attribution is needed amidst various end users who can train new models using their own datasets.
This result is in alignment with the way we constructed our definition of fingerprints: since they are defined as the differences between the true data manifold and the generated samples, it has the effect of ``subtracting away'' the fingerprints' dependence on the choice of training datasets, thereby improving the cross-dataset generalizability.

%% file: tbls/tbl-exp-result-2-layer-types.tex
\begin{table}[t]

\setlength{\tabcolsep}{.5em}
\centering
\small
\begin{tabular}{@{}lc ccccc c@{}}
\toprule
     & (NMI$\uparrow$) & Layer &  Types &   &   & Optim. \\
\cline{2-6}\cline{6-7}\vspace{-1em} \\
Method &  \textbf{\textit{Up}} & NL & Norm & Down & Skip & \textbf{\textit{Loss}}  \\

\toprule
    $\oursrgb{}$   & 0.625& 0.453 & 0.647 & 0.432 & 0.541 & 0.563 \\
    $\oursfreq{}$  & 0.654 & 0.354 & 0.534 & 0.692 & 0.317 & 0.631 \\
    $\ourssup{}$   & 0.613 & 0.452 & 0.481& 0.546 & 0.434 & 0.677\\
    $\oursunsup{}$ & 0.680 & 0.477 & 0.465 &0.615 & 0.357& 0.573 \\
\hdashline
    Average & \textbf{0.643} & 0.434 & 0.465 &0.532 & 0.571 & \underline{0.611} \\
\bottomrule
\end{tabular}
\caption{\small
\textbf{Clustering structure of fingerprints.}
Types of upsampling and loss best align with the clustering. 
\textbf{NL}: Non-linearity
}
\label{tab:exp:classify-artifacts-results}
\vspace{-1em}
\end{table}

%% file: rqs/rq2-clustering-structure-of-artifacts.tex
Lastly, we study the structure of GM fingerprints by studying the alignment of their clustering pattern to  the hyperparameters used in the design of generative models (\eg type of sampling layers and loss functions).
Table ~\ref{tab:exp:classify-artifacts-results} reports the alignment results.
Overall, we observe that upsampling and loss types best match the clustering behavior of the artifacts,  experimentally confirming the general intuition about the sources of limitations in generative models and supporting the utility of our definitions in studying the model behaviors. 
See full discussion in Appendix~\ref{subsec:exp:characterize}.

%% file: sec/05-conclusion.tex
Our work addresses the emerging problem of differentiating generative models and attributing generated samples to their proper source models. 
We study this problem in a principled way by proposing formal definitions of artifacts and fingerprints of generative models, which has been missing in the literature. 
We further provide a theoretical justification of our proposed definitions in relation to two key metrics on generative models (Precision and Recall, and IPMs), and
demonstrate their practical usefulness in differentiating a large array of state-of-the-art models, of various types (GAN, VAE, Flow and Score-based).
Our method outperforms existing methods on model attribution, generalizes better across datasets, 
and learns a feature space effective for differentiating generative models.
We believe our definitions will lay an important step towards formalizing the characteristics of generative models
and prepare for their integration into our society by 
helping to develop more effective attribution methods.

%% file: sec/06-appendix.tex
\clearpage
\newpage
\appendix
\section{Dataset Creation} \label{appendix:our-datasets}
As  discussed in Sec.~\ref{sec:rel-work}, existing datasets designed for the binary discrimination of real vs. synthetic samples are not suitable for the task of model attribution (\ie discriminating among multiple different generative models) in two aspects:
(i) the diversity of generative models (GMs) is limited, and
(ii) the variability of the models' training datasets makes the study of model fingerprint -- independent of their training datasets -- difficult.  
Rather, a proper benchmark dataset  for model attribution  should satisfy the following desiderata:

\begin{enumerate}
\item It should include GMs from various families, covering VAEs, GANs, Flows and Score-based (\ie Diffusion) models
\item It should contain state-of-the-art models, in addition to the more standard models that existing works have focused on (\eg ProGAN, CycleGAN, StyleGAN)
\item The generative models in the dataset should be trained on the same training set in order for the analysis on the fingerprint features to be directly attributed to the 
characteristics of the generative models, without confounding effects from the variability in the models' training datasets. 
\end{enumerate}
To this end, we designed three new datasets (\gmcifar{}, \gmceleba{}, \gmchq{} and \gmffhq{}; See Tbl.~\ref{tbl:exp-datasets}) that carefully satisfy these three desiderata.
\gmcifar{} contains images from generative models trained on CIFAR-10~\cite{Krizhevsky2009LearningML}.
\gmceleba{} contains images from generative models trained on CelebA~\cite{liu2015faceattributes},
\gmchq{} from models trained on CelebA-HQ (256)~\cite{Karras2018ProgressiveGO},
and \gmffhq{} from models trained on FFHQ (256)~\cite{Karras2018ProgressiveGO}.
To complement the existing datasets, our datasets include  GMs that achieve state-of-the-art results on unconditional image synthesis, such as DDGAN~\cite{xiao2022TACKLINGGENERATIVELEARNING} and StyleSwin~\cite{zhang2022StyleSwinTransformerbasedGANa} for GAN, NVAE~\cite{vahdat2020NVAEDeepHierarchical} and Efficient-VDVAE~\cite{hazami2022EfficientVDVAELessMore} for VAE, and  LDM~\cite{rombach2022HighResolutionImageSynthesisb} and LSGM \cite{vahdat2021ScorebasedGenerativeModeling} for diffusion models.

\subsection{Details on dataset creation}

\noindent{}\textbf{\gmceleba ~dataset}
We construct a dataset of real and GM-generated images by collecting real images from the original CelebA (image-aligned and resized to 64x64)~\cite{liu2015faceattributes} and generating 100k samples from GMs trained on CelebA-64.
We consider 4 VAE models, 5 GAN models, 1 Flow and 1 Score-based model, based on the best availability of the released code and model checkpoints. 
We trained the models on our own when no official pretrained model was released. See Tab.~\ref{tbl:exp-datasets} for the list of GMs used for this dataset.

\noindent{}\textbf{\gmchq{}~dataset}
To study the fingerprints of more advanced generative models, 
we collect samples from state-of-the-art models such as NVAE~\cite{vahdat2020NVAEDeepHierarchical}, Efficient VDVAE~\cite{hazami2022EfficientVDVAELessMore}, VQ-GAN~\cite{esser2021TamingTransformersHighResolutiona}, StyleGAN2~\cite{karras2020AnalyzingImprovingImagea}, Denoising Diffusion GAN (DDGAN)~\cite{xiao2022TACKLINGGENERATIVELEARNING}, DDPM++~\cite{ho2020DenoisingDiffusionProbabilistic}, NCSN++~\cite{song2023ScoreBasedGenerativeModeling} and Latent Score-based Generative Model (LSGM)~\cite{vahdat2021ScorebasedGenerativeModeling}. 
All the models are trained on CelebA-HQ 256~\cite{Karras2018ProgressiveGO}.
From each model, we collect 100k samples. 
See Tab.~\ref{tbl:exp-datasets} for the full list of GMs and the supplementary materials for details on the sampling procedure from each generative model.
d

Fig.~\ref{fig:our-gm256-gans-thumbnails}, Fig.~\ref{fig:our-gm256-vaes-thumbnails} and Fig.~\ref{fig:our-gm256-scores-thumbnails} show samples from our \gmchq{} dataset. The images are randomly sampled from each GMs following the process detailed in each work or codebase. 

\section{Details on baseline fingerprinting methods} 
\label{appendix:baselines}
Tab.~\ref{tbl:baselines-datasets} summarizes baseline fingerprinting methods that we compared against our definitions proposed in Sec.~\ref{sec:exps}.
\input{tbls/tbl-baselines} 

\section{Feature space analysis} \label{appendix:fdr}
\subsection{Fréchet Distance Ratio (FDR)}
We measure the separability of a fingerprint feature space using the ratio of Fréchet Distance. 
This measure was also used in Yu et al.~\cite{yu2019AttributingFakeImages} to evaluate the learned feature space for GAN fingerprints. 
In our work, we use it to evaluate fingerprints in a more generalized sense in that they are to identify more diverse set of GMs (not just GANs) including many state-of-the-art models.

FDR is computed as the ratio of inter-class and intra-class Fréchet Distance~\cite{dowson1982frechet}: 
\begin{equation}
    FDR = \frac{ \textrm{inter-class FD} }{\textrm{intra-class FD}  }
\end{equation}

\noindent\textbf{Intra-class FD} aims to capture the average tightness of a feature distribution per class, and can be measured as the FD between two disjoint sets of images in the same class.
As in Yu et al.~\cite{yu2019AttributingFakeImages}, we split, for each class, the fingerprint features into two disjoint sets of equal size, compute their Fréchet Distance, and then average it over each class. 

\noindent\textbf{Inter-class FD} aims to capture the average distance between feature distributions of different classes. 
To compute this distance, we measure the FD between two feature sets from different classes and take the average over every possible pair of (different) classes.

 \input{tbls/tbl-exp-result-2-characterize} 
 \section{Experiment: characterization of generative models} \label{subsec:exp:characterize} 
 \input{rqs/rq2-classification-of-artifacts}


\section{Visualization: artifacts of generative models}
We visualize more examples of artifacts of generative models in \gmceleba{} and \gmchq{}, computed under our proposed definition in Sec.~\ref{sec:our-defns}. Fig.~\ref{fig:suppl-viz-art-rgb-triplets-gm64} shows the triplets of (generated images ($x_G$), its closest point to the data manifold in RGB ($x^{\star}$) and the artifact ($a$)).
Fig.~\ref{fig:suppl-viz-art-fft-triplets-gm256} visualize the artifacts in frequency domain from the GM-CHQ dataset. 

\subsection{Artifacts in RGB space (\gmceleba{})} 
\input{tbls/fig-suppl-viz-art-rgb-gm64}

\input{tbls/fig-suppl-viz-art-fft-gm256}

\input{tbls/fig-our-gm256-thumbnails}

%% file: tbls/tbl-baselines.tex
\begin{table*}[t] 
\begin{center}
\begin{adjustbox}{width=1.\textwidth}
%
\begin{tabular}{c|c|c|c|c|c}
\toprule
Paper & Input domain & Representation & Classifiers & Metric(best) & Datasets \\
\toprule
%
McCloskey18~\cite{mccloskey2018DetectingGANgeneratedImagerya}  &  
RGB &
\makecell{Histogram of saturated, \\ under-exposed pixels}  & 
SVM & 
AUC (0.7) & 
NIST MFC2018 
\\
%
\hline
Nataraj19~\cite{nataraj2019DetectingGANGenerateda}  &  
RGB &
\makecell{Co-occurrence matrix \\ of pixels} & 
CNN & 
EER (12.3\%) & 
100k-Faces (StyleGAN)\\
\hline \hline
%
Durall20~\cite{durall2020WatchYourUpConvolution}  & 
Freq. &
\makecell{1D power spectrum \\(azimuthal integral) }& 
SVM & 
\makecell{Binary Acc \\(96\%)} & 
\makecell{Own \\ (DCGAN, DRAGAN, \\ SGAN, WGAN-gp)} 
\\
%
\hline
Dzanic20~\cite{dzanic2020FourierSpectrumDiscrepancies} &  
 Freq.&
 \makecell{Fourier spectrum \\ (norm. by DC gain)}& 
 KNN & 
\makecell{Binary Acc \\ (99.2\%)} & 
 \makecell{Own \\ (StyleGAN,StyleGAN2,\\PGGAN,VQ-VAE2,ALAE)}
\\
%
\hline
Wang20~\cite{wang2020CNNGeneratedImagesAre}  &  
Freq. &
2D average spectra & 
\makecell{CNN} &
\makecell{LOMO, Binary Acc \\ (84.7\%)} & 
Own (10 GANs)
\\
\hline \hline
Marra18~\cite{marra2018DetectionGANGeneratedFake}  &  
Learned &
Supervised &
\makecell{Pretrained CNN + Finetuned\\ (Inception-v3/XceptionNet) } & 
\makecell{LOMO\footnote{Leave-One-Manipulation-Out (Sec.3 protocol \cite{marra2018DetectionGANGeneratedFake}}, Binary Acc \\ (94.49\%)} & 
\makecell{Own \\(Real, CycleGAN per category)} 
\\
%
Marra19~\cite{marra2019IncrementalLearningDetectiona}  &  
Learned &
Supervised & 
\makecell{CNN + IL} & 
\makecell{Binary Acc \\ (99.3\%)} & 
\makecell{Own \\(4 GANs, 1 Flow)} 
\\
%
\hline
Yu19~\cite{yu2019AttributingFakeImages}  &  
Learned &
Supervised & 
CNN & 
\makecell{Multi Acc \\ (98.58\%)} & 
\makecell{Own \\(ProGAN, SNGAN \\ CramerGAN, MMDGAN)}
\\


\bottomrule
\end{tabular}
\end{adjustbox}
\caption{\textbf{Features and datasets used in the baseline methods} 
}
\label{tbl:baselines-datasets}
\end{center}
\vspace{-0.5em}

\end{table*}


%% file: tbls/tbl-exp-result-2-characterize.tex
\begin{table*}[t]

\setlength{\tabcolsep}{.5em}
    \centering
      \begin{tabular}{@{}lc ccccc c@{}}
           &&  &  & Model Params.  (NMI $\uparrow$)  &  &   & Optim. Params \\
	      \cline{1-8}\vspace{-1em} \\
    Methods &&  \textbf{\textit{Upsampling}} & Non-linearity & Normalization & Downsampling & Use skip & \textbf{\textit{Loss Type}}  \\
	\cline{1-1}\cline{2-8}\vspace{-1em}\\
	\cline{1-1}\cline{2-8}\vspace{-1em} \\
    $\oursrgb{}$   && 0.625& 0.453 & 0.647 & 0.432 & 0.541 & 0.563 \\
    $\oursfreq{}$  && 0.654 & 0.354 & 0.534 & 0.692 & 0.317 & 0.631 \\
    $\ourssup{}$   && 0.613 & 0.452 & 0.481& 0.546 & 0.434 & 0.677\\
    $\oursunsup{}$ && 0.680 & 0.477 & 0.465 &0.615 & 0.357& 0.573 \\
        \cline{1-1}\cline{2-8} \vspace{-1em} \\
    Average && \textbf{0.643} & 0.434 & 0.465 &0.532 & 0.571 & \underline{0.611} \\

    \cline{1-1}\cline{2-8}\vspace{-1em}\\
    \end{tabular}
    \caption{
    {\bf Clustering structure in \gmchq{}.} We measure the alignment of clustering in Normalized Mutual Information (NMI) on our feature spaces (using RGB, Frequency, Supervised-learning (SL), Self-supervised learning (SSL) representations to clusterings based on model design parameters (\eg type of upsampling, type of non-linearity in the last layer, type of normalization later, type of loss function). 
    NMI is bounded to [0,1]. Higher index indicates closer agreement between two cluster assignments.
     }
    \label{tab:exp:classify-artifacts-results-2}
\end{table*}

%% file: rqs/rq2-classification-of-artifacts.tex
We further study the clustering structure of the set of GM artifacts and explore if it is possible to relate the clustering patterns to the hyperparameters that govern the model design of  generative models, such as the type of sampling layers and the type of loss functions.
To do so, we take insights from the experiments in \cite{Asnani2021ReverseEO,durall2020WatchYourUpConvolution, dzanic2020FourierSpectrumDiscrepancies}, and group the model hyperparameters into the following categories: 
Type of upsampling,
Type of non-linearity in the last layer,
Type of normalization,
Use of downsampling,
Use of skip connection, 
and Type of loss function.
For example, we categorize the loss functions in our datasets into three types (likelihood-based (VAEs), implicit density matching (GANs), and score-matching (Score-based models)), 
and the type of non-linearity in the last layer into ReLU, Tanh, and Sigmoid.
See the supplementary for more details on our categorization of the hyperparameters and specific values each GM in our datasets take. 

\noindent\textbf{Metric.} 
We use Normalized Mutual Information (NMI)~\cite{mcdaid2011normalized} to measure the clustering alignment between the clustering in a fingerprint representation ($\mathcal{C}_{f}$) and the clustering on the assignment of a model design choice (\eg type of upsampling operation) as the ground-truth cluster labels ($\mathcal{C}_{h}$). 
For instance, to measure how well the clustering in a fingerprint space coincides with the clusters according to the type of loss function,
we set as $\mathcal{C}_{h}$ the result of clustering datapoints based on the type of their source GM's loss function.
If the loss type is a proper criterion to categorize different generative models, the two clusterings ($\mathcal{C}_{f}$ based on the fingerprint representations and $\mathcal{C}_{h}$ on the labels of loss type) will have a strong agreement, 
and their clustering index will be high.

\noindent\textbf{Results.}
Table ~\ref{tab:exp:classify-artifacts-results} reports NMI between a feature space and each category of model hyperparameters, reflecting which criterion in grouping the generative models (\eg the type of upsampling vs. the type of non-linearity vs. the type of loss function) agrees well with the grouping in the fingerprint representation space.  
Note that NMI is bounded to $[0,1]$, and a higher index indicates a closer agreement between two cluster assignments. 
The last row  ($\emph{Avg}\textsubscript{ours}$) shows the NMI averaged over our methods in RGB, frequency, supervised-learning based and unsupervised-learning based representation space. 
First of all, we observe that the clustering on our fingerprint space aligns the best with the clustering by the GMs' upsampling types and loss types. 
In other words, our result suggests that the two hyperparameters (Type of upsampling and Type of loss function) show the most similar clustering patterns with our fingerprint representations. 

The high NMI for the type of upsampling supports previous experiments that identified the upsampling operation of generator networks as a cause of the high-frequency discrepancies in the GM-generated images~\cite{chandrasegaran2021CloserLookFourier, dzanic2020FourierSpectrumDiscrepancies, durall2020WatchYourUpConvolution}.
Additionally, the high NMI for the type of loss function confirms the general consensus in the research community that the training objective of a generative model is one of the key factors that affect their characteristics.

Therefore, our findings confirms the general intuition in the research community about distinct sources of limitations in generative models and shows the utility of our definitions.

%% file: tbls/fig-suppl-viz-art-rgb-gm64.tex


\begin{figure}[h]
  \centering
  \includegraphics[
  height=0.45\textheight, 
  keepaspectratio]
  {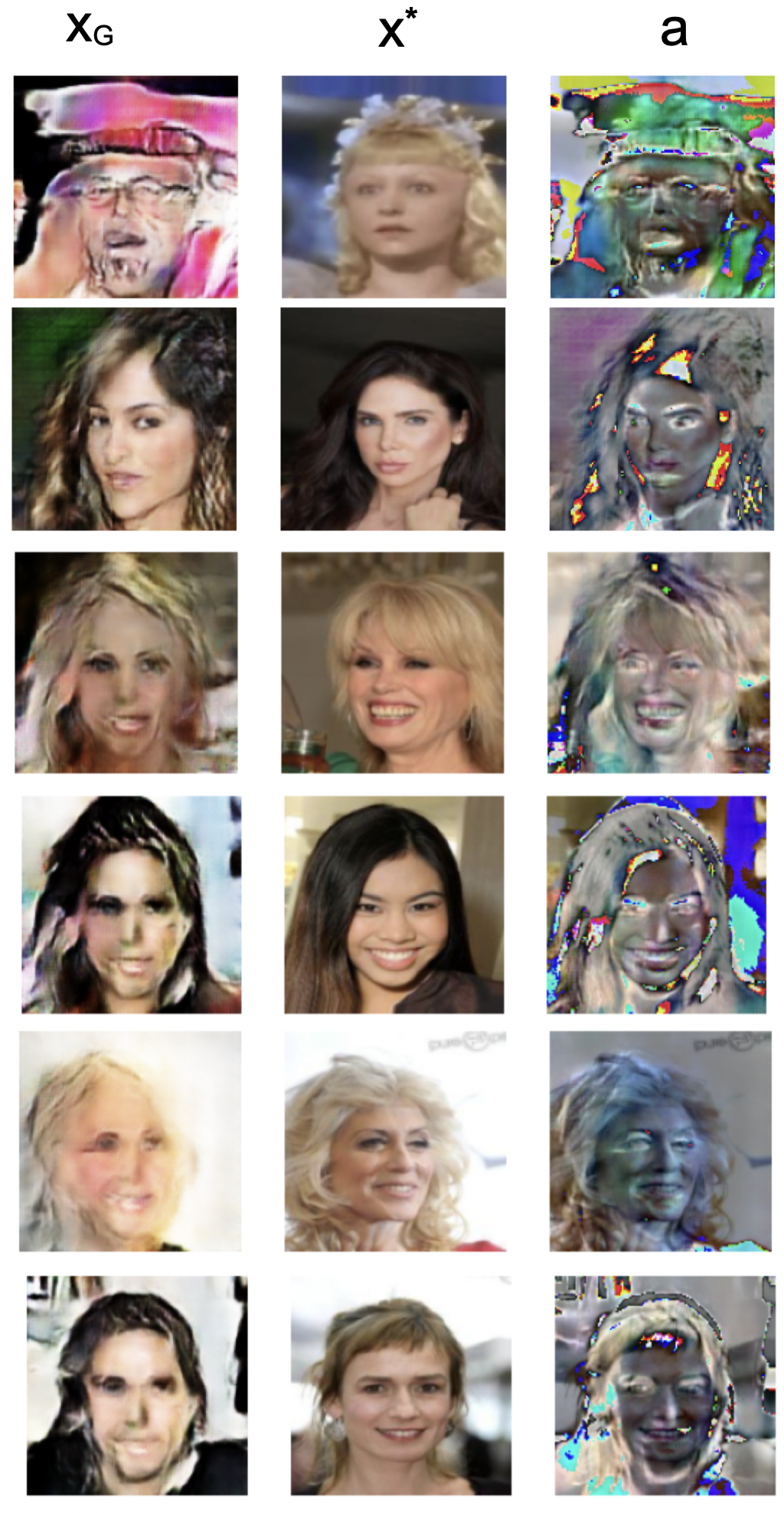}

  \caption{\textbf{Visualization of artifacts in the RGB space (\gmceleba{}).}
  Each column corresponds to the generated images ($x_G$), their closest points on the data manifold ($x^{\star}$), and the artifacts ($a$).
  Each artifact is computed as the different between $x^{\star}$ and $x_G$ following the definition and algorithm in 
    Sec.~\ref{sec:our-defns}. 
  }
  \label{fig:suppl-viz-art-rgb-triplets-gm64}
\end{figure}

%% file: tbls/fig-suppl-viz-art-fft-gm256.tex
\begin{figure*}[ht]
  \centering
  \begin{subfigure}[b]{0.475\textwidth} 
  \includegraphics[
  width=\textwidth,
  height=0.8\textheight,
  keepaspectratio
  ]{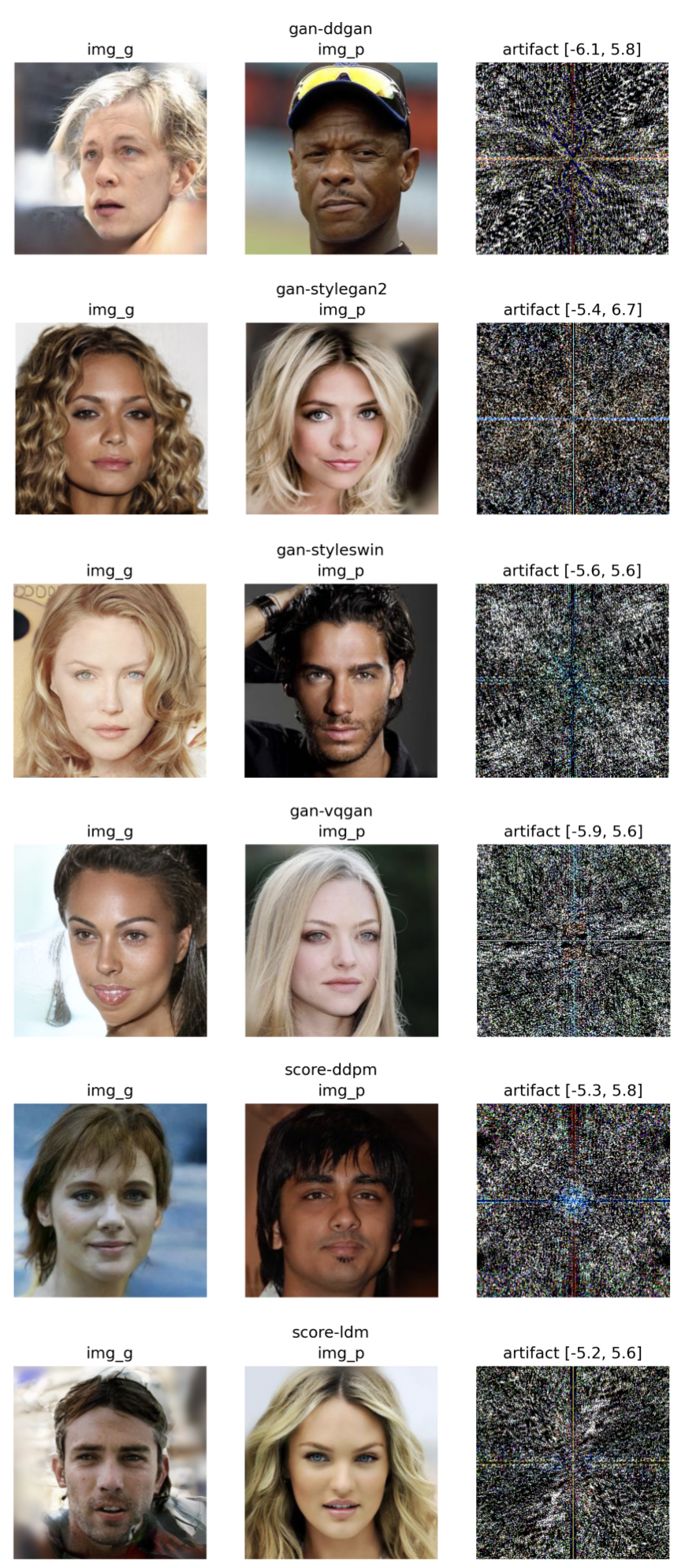}
\end{subfigure}
\hfill
\begin{subfigure}[b]{0.475\textwidth}
  \includegraphics[
  width=\textwidth,
  height=0.8\textheight,
  keepaspectratio
  ]{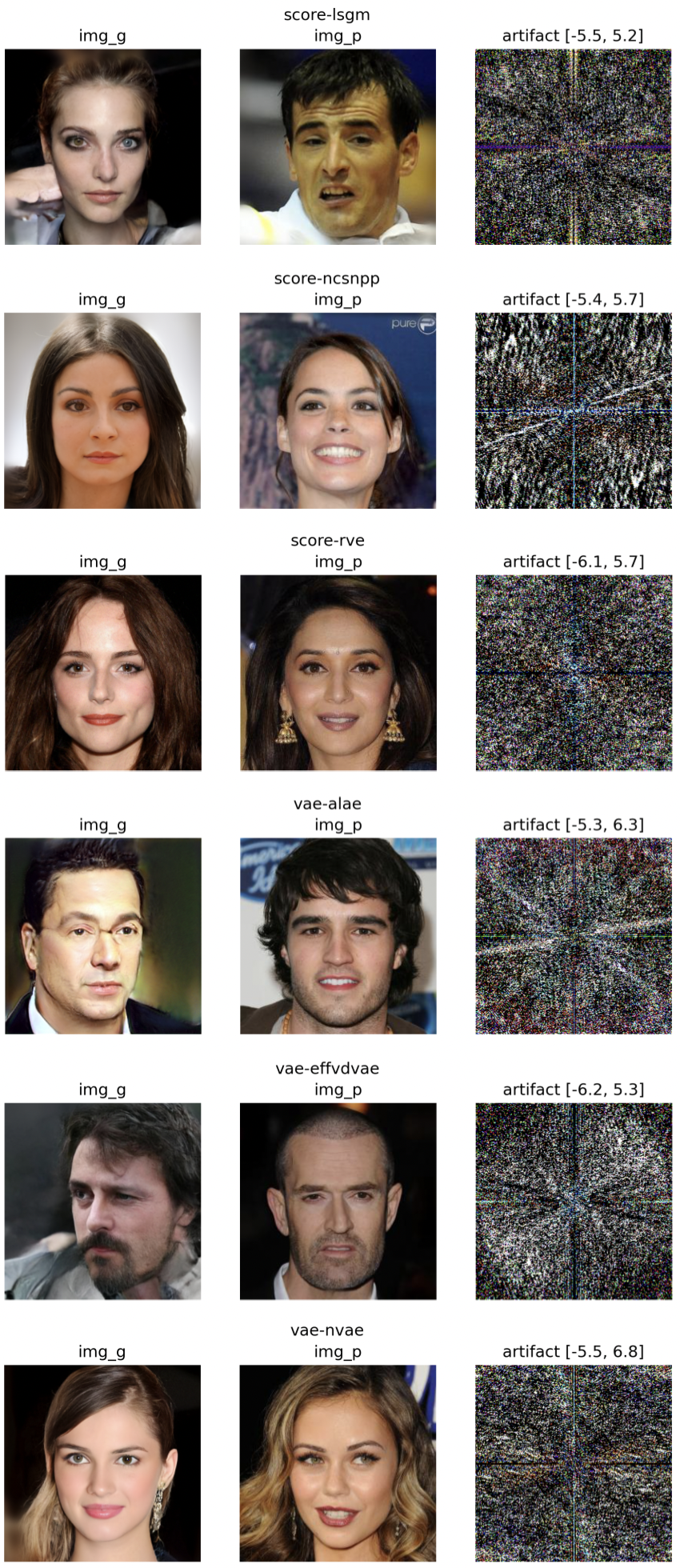}
  \end{subfigure}
  \caption{\textbf{Visualization of artifacts in the frequency space (\gmchq{)}}. We show some examples of triplets (model-generated image ($\textrm{img}_g$), closest point on the data manifold ($\textrm{img}_p$), artifact) from \gmchq{} dataset by computing artifacts (as defined in Sec.~\ref{sec:our-defns}) in frequency domain. $\textrm{img}_p$ is the point on the real data manifold that is closest to $\textrm{img}_g$ in the frequency domain. Artifact is computed as the different between the two points, $\textrm{img}_g$ and $\textrm{img}_p$, after applying channelwise-FFT.}
    \label{fig:suppl-viz-art-fft-triplets-gm256}

\end{figure*}

%% file: tbls/fig-our-gm256-thumbnails.tex
\begin{figure*}[t]
\centering
\begin{subfigure}[b]{0.475\textwidth} 
    \centering
    \includegraphics[width=\textwidth]{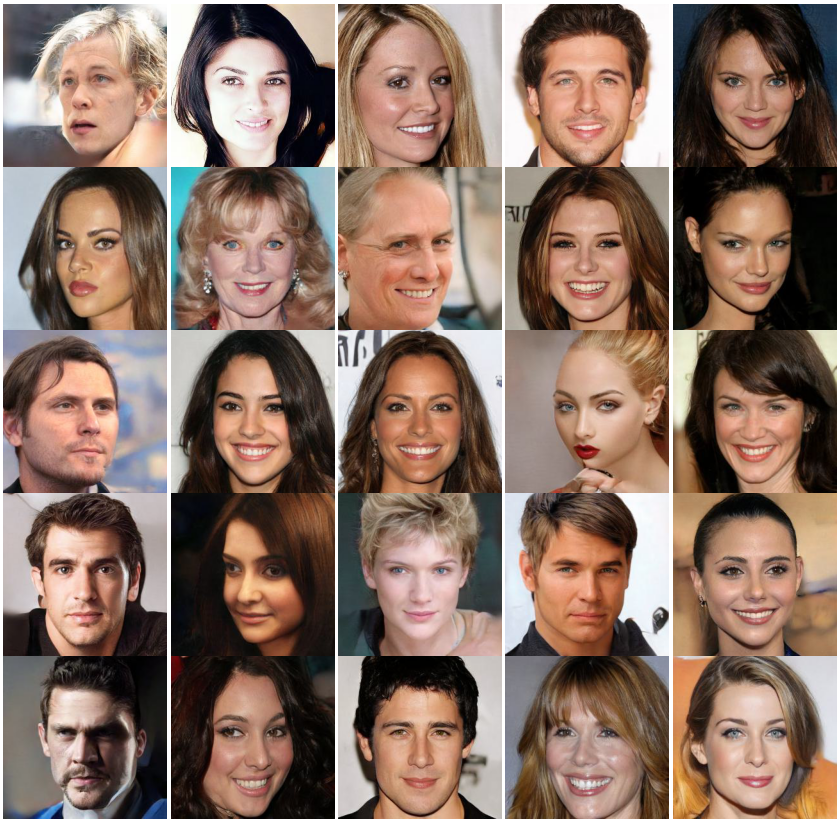}
    \caption{DDGAN \cite{xiao2022TACKLINGGENERATIVELEARNING}}
\end{subfigure}
\hfill
\begin{subfigure}[b]{0.475\textwidth}
    \centering
    \includegraphics[width=\textwidth]{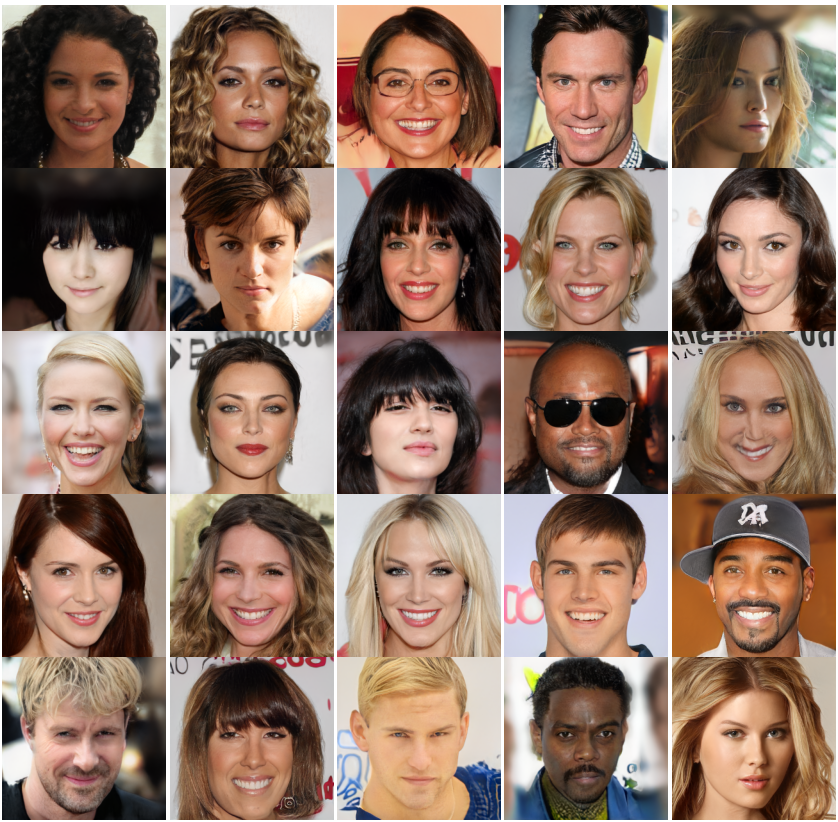} 
    \caption{StyleGAN2 \cite{karras2020AnalyzingImprovingImagea}}
\end{subfigure}
\vskip\baselineskip
\begin{subfigure}[b]{0.475\textwidth}
    \centering
    \includegraphics[width=\textwidth]{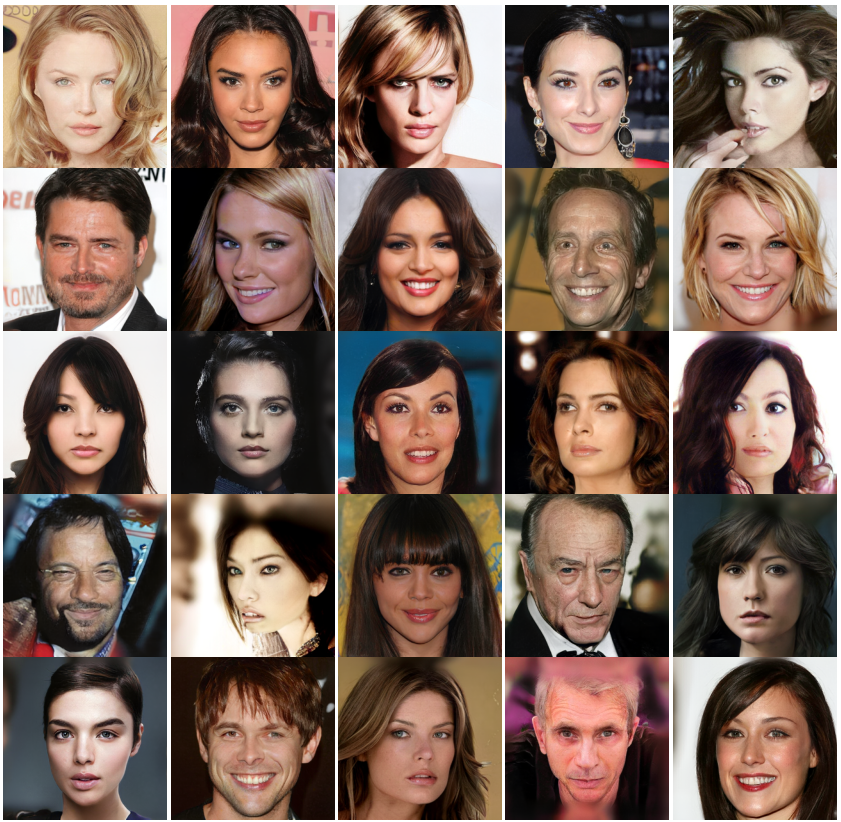}
    \caption{StyleSwin \cite{zhang2022StyleSwinTransformerbasedGANa}}
\end{subfigure}
\hfill
\begin{subfigure}[b]{0.475\textwidth}
    \centering
    \includegraphics[width=\textwidth]{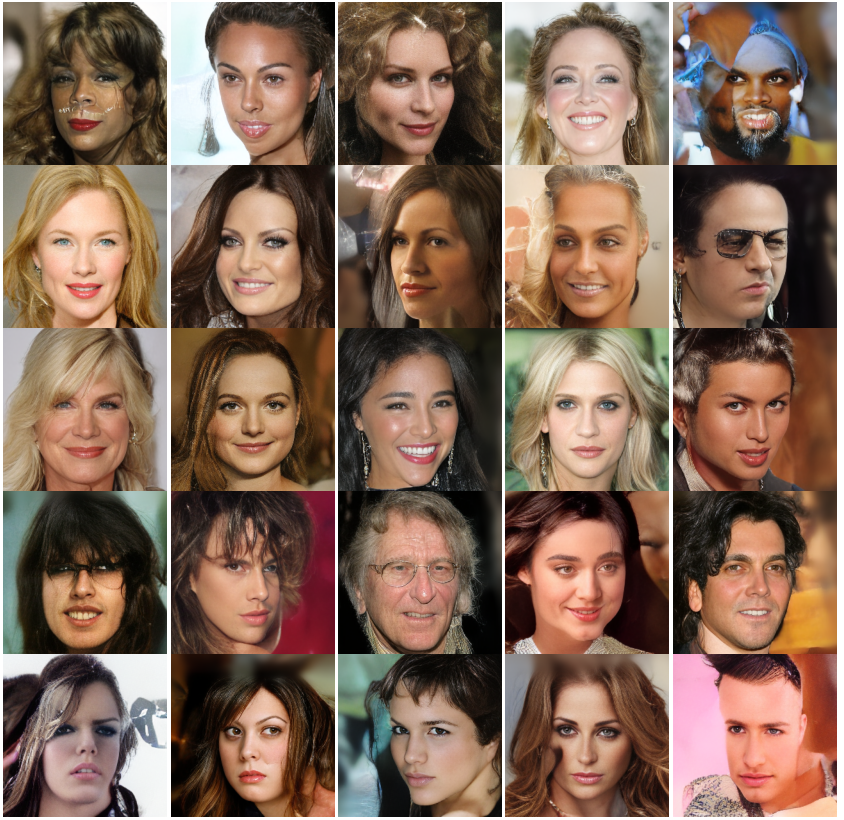}
    \caption{VQ-GAN \cite{esser2021TamingTransformersHighResolutiona}}
\end{subfigure}
\caption{\textbf{Samples from GAN models in \gmchq. }}
  \label{fig:our-gm256-gans-thumbnails}
\end{figure*}

\begin{figure*}[t]
\centering
\begin{subfigure}[b]{0.475\textwidth} 
    \centering
    \includegraphics[width=\textwidth]{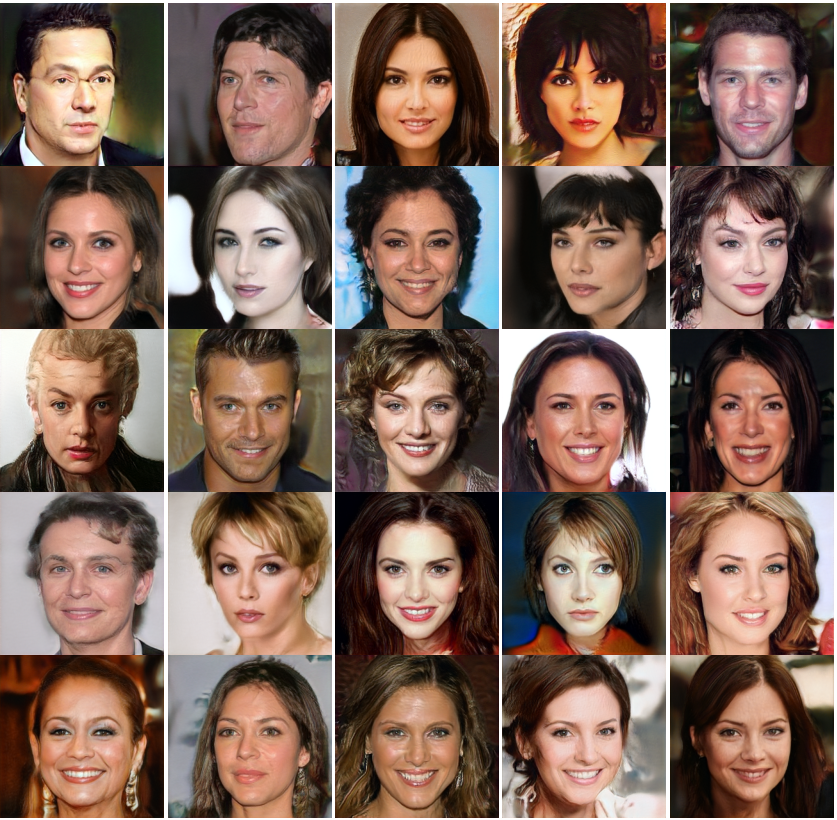}
    \caption{StyleALAE \cite{pidhorskyi2020AdversarialLatentAutoencoders} }
\end{subfigure}
\hfill
\begin{subfigure}[b]{0.475\textwidth}
    \centering
    \includegraphics[width=\textwidth]{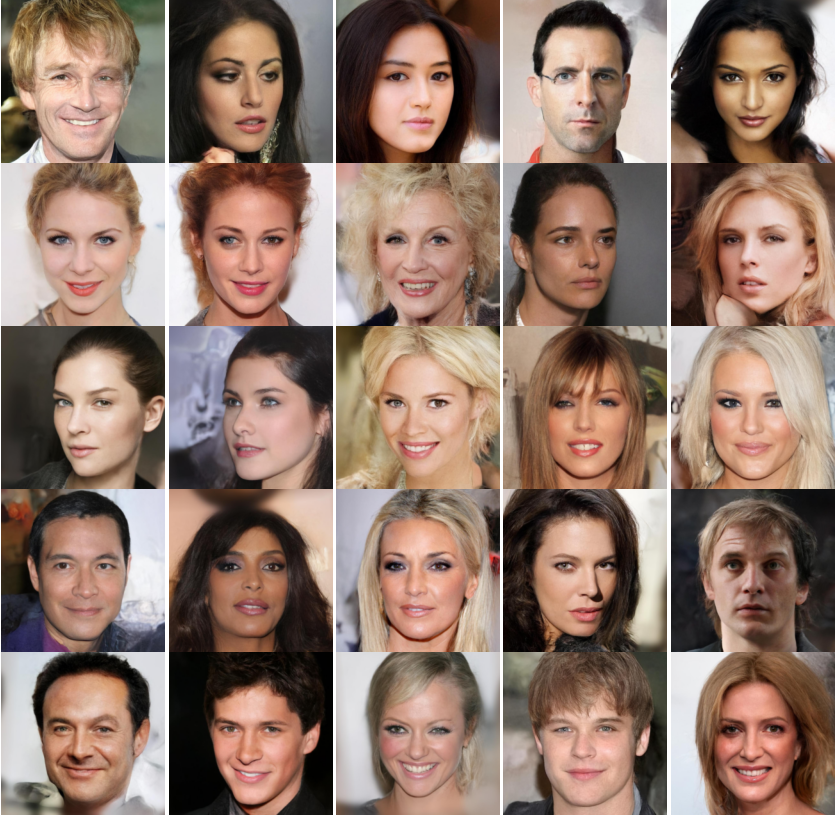} 
    \caption{Efficient VDVAE~\cite{hazami2022EfficientVDVAELessMore}}
\end{subfigure}
\vskip\baselineskip
\begin{subfigure}[b]{0.475\textwidth}
    \centering
    \includegraphics[width=\textwidth]{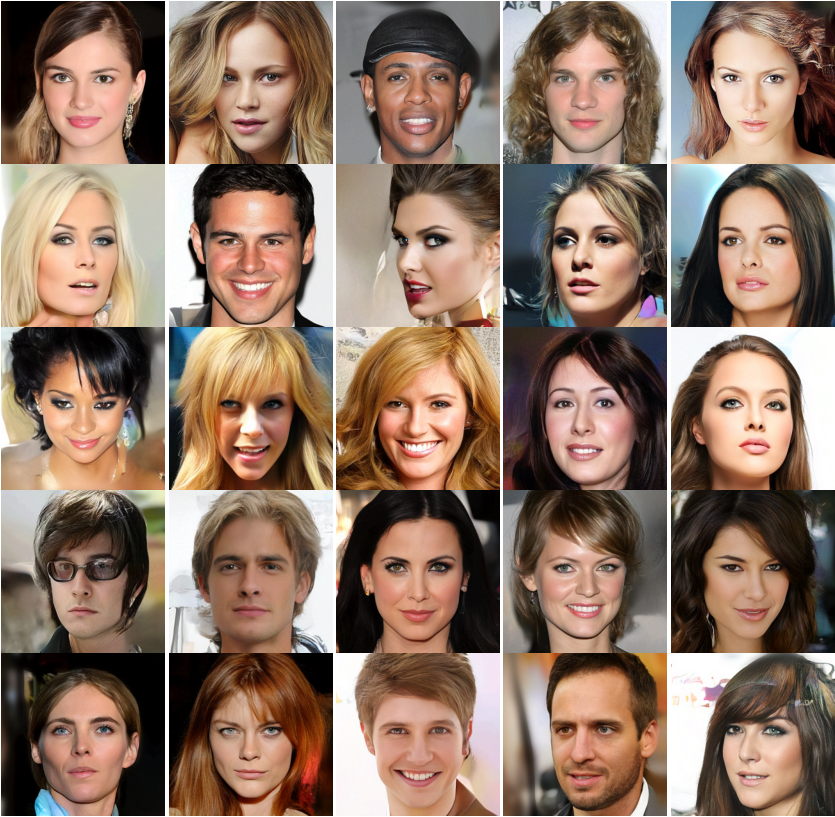}
    \caption{NVAE~\cite{vahdat2020NVAEDeepHierarchical}}
\end{subfigure}
\hfill
\begin{subfigure}[b]{0.475\textwidth}
    \centering
    \includegraphics[width=\textwidth]{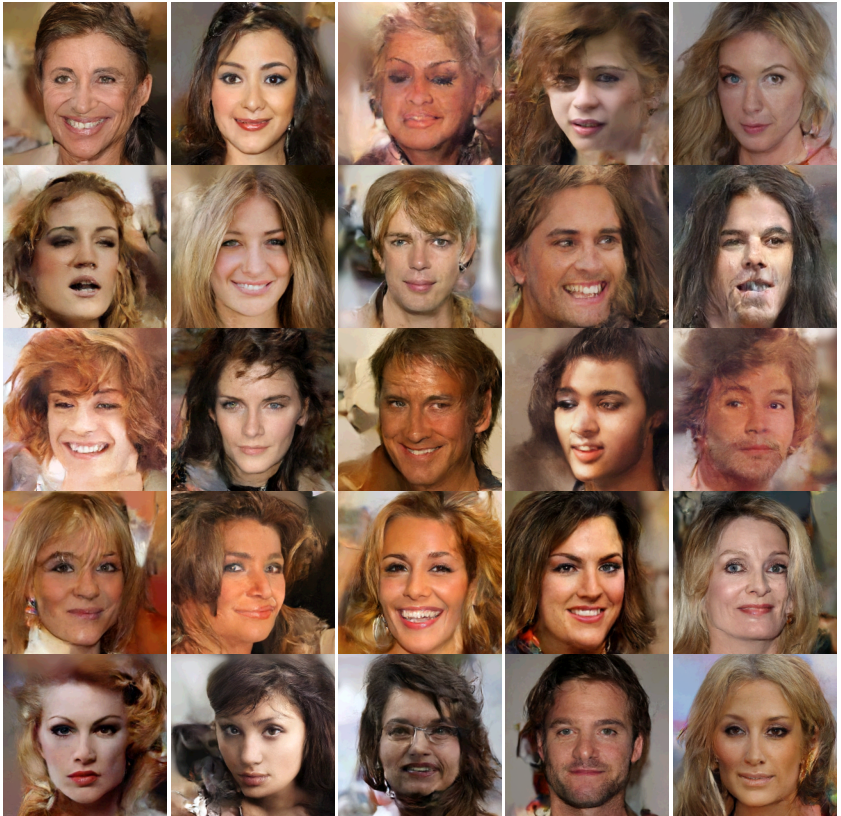}
    \caption{ VAEBM~\cite{xiao2022VAEBMSymbiosisVariational}}
\end{subfigure}
\caption{\bf Samples from VAE models in \gmchq. }
\label{fig:our-gm256-vaes-thumbnails}
\end{figure*}

\begin{figure*}[t]
\centering
\begin{subfigure}[b]{0.475\textwidth} 
    \centering
    \includegraphics[width=\textwidth]{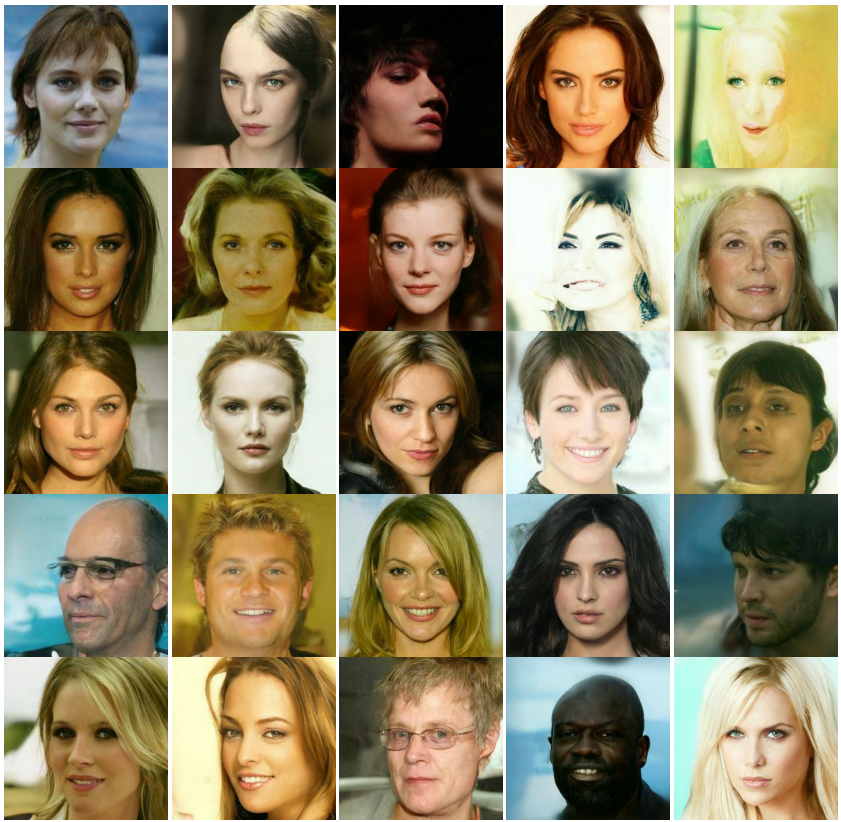}
    \caption{DDPM \cite{ho2020DenoisingDiffusionProbabilistic} }
\end{subfigure}
\hfill
\begin{subfigure}[b]{0.475\textwidth}
    \centering
    \includegraphics[width=\textwidth]{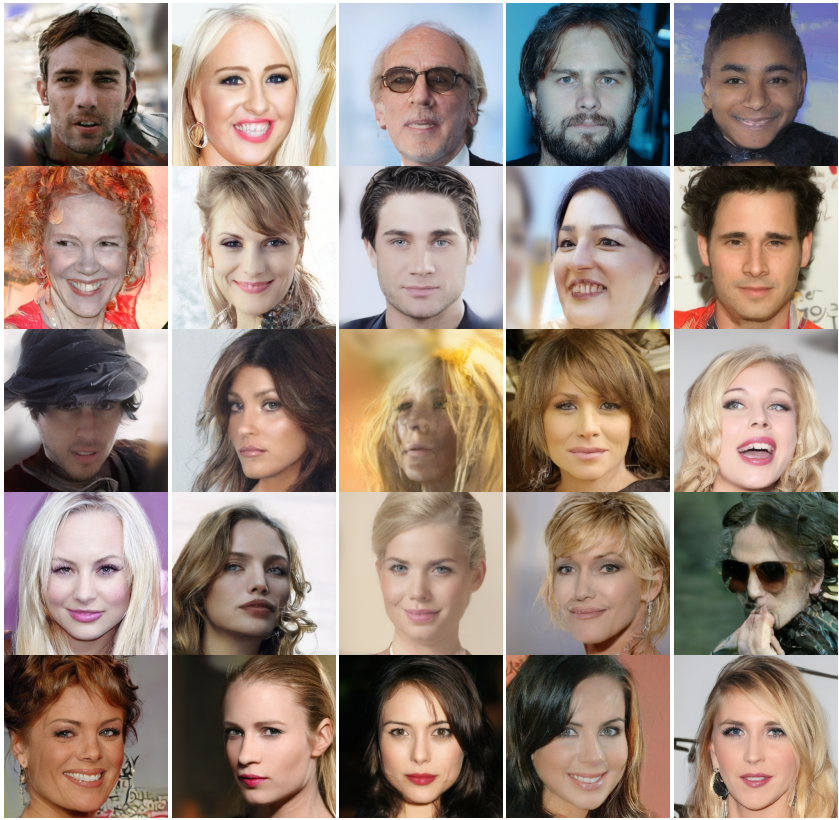} 
    \caption{LDM \cite{rombach2022HighResolutionImageSynthesisb}}
\end{subfigure}
\vskip\baselineskip
\begin{subfigure}[b]{0.475\textwidth}
    \centering
    \includegraphics[width=\textwidth]{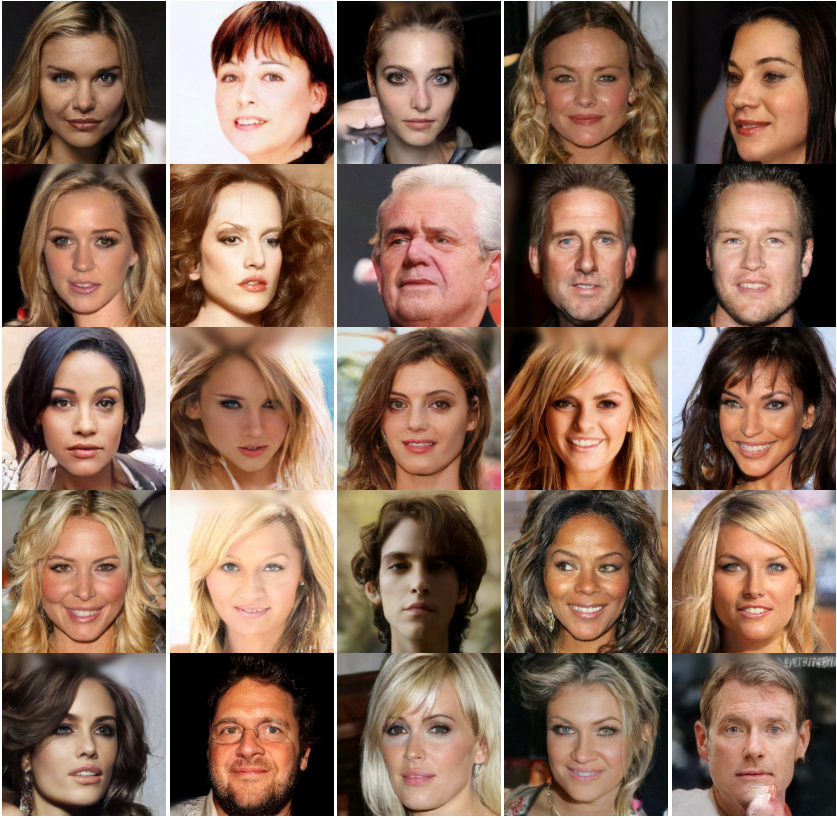}
    \caption{LSGM \cite{vahdat2021ScorebasedGenerativeModeling}}
\end{subfigure}
\hfill
\begin{subfigure}[b]{0.475\textwidth}
    \centering
    \includegraphics[width=\textwidth]{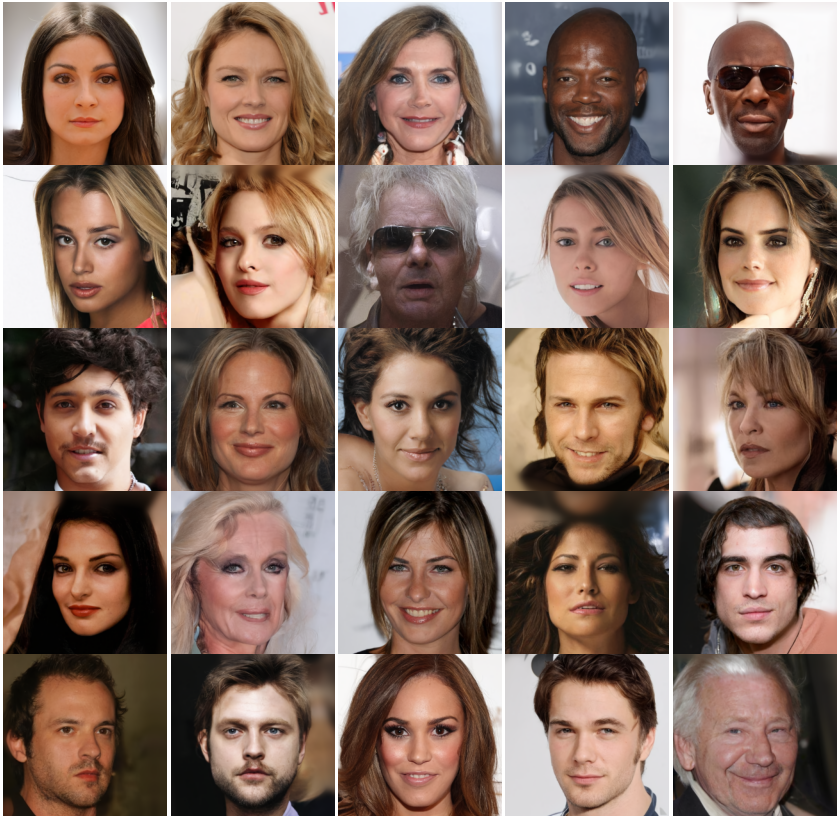}
    \caption{NCSN++ \cite{song2023ScoreBasedGenerativeModeling} }
\end{subfigure}
\caption{\bf Samples from score-based (a.k.a. diffusion) models in \gmchq. }
  \label{fig:our-gm256-scores-thumbnails}
\end{figure*}